\documentclass[letterpaper, 10 pt, conference]{ieeeconf}  

\IEEEoverridecommandlockouts                              

\usepackage{tabularx,booktabs}
\usepackage[flushleft]{threeparttable}
\usepackage{nameref}
\usepackage{amsmath}
\usepackage[utf8]{inputenc}
\usepackage{booktabs}
\usepackage{cite}
\usepackage{graphicx}
\usepackage{color}
\usepackage{url}
\usepackage{cite}
\usepackage{comment}
\usepackage{amsfonts}
\usepackage{hyperref}
\usepackage[dvipsnames]{xcolor}

\usepackage{multicol}
\usepackage[per-mode=symbol]{siunitx}
\usepackage{lscape}
\usepackage{pifont}
\usepackage{xcolor}
\usepackage{makecell}
\usepackage{soul}
\usepackage{graphicx}
 \usepackage{amsmath}
 \usepackage{amsfonts}
 \usepackage{mathtools}
 \usepackage{bm}
 \usepackage{array}
 \usepackage{multirow}
 \usepackage{adjustbox}
 \usepackage{tikz}
 \usetikzlibrary{calc}

\usepackage{xurl}
\hypersetup{
    colorlinks=true,
    linkcolor=RoyalBlue,
    filecolor=DarkSlateGray,      
    urlcolor=RoyalBlue,
    pdftitle={Overleaf Example},
    pdfpagemode=FullScreen,
    }

\usepackage[per-mode=symbol]{siunitx}
\pdfoutput=1
\usepackage{tablefootnote}
\DeclareSIUnit{\fps}{ fps }

\title{\LARGE \bf
GuideTWSI: A Diverse Tactile Walking Surface Indicator Dataset from Synthetic and Real-World Images for Blind and Low-Vision Navigation
}

\author{Hochul Hwang$^{1}$, Soowan Yang$^{2}$, Anh N. H. Nguyen$^{1}$, Parth Goel$^{1}$, Krisha Adhikari$^{1}$, \\ Sunghoon I. Lee$^{1}$, Joydeep Biswas$^{3}$, Nicholas A. Giudice$^{4}$, and Donghyun Kim$^{\dagger}$
\thanks{$^{1}$ University of Massachusetts Amherst}%
\thanks{$^{2}$ Daegu Gyeongbuk Institute of Science and Technology }%
\thanks{$^{3}$ The University of Texas at Austin}%
\thanks{$^{4}$ University of Maine}%
\thanks{$^{\dagger}$University of Massachusetts Amherst, 140 Governors Dr, U.S. {\tt\small donghyunkim@cs.umass.edu}}
}

\begin{document}

\maketitle
\thispagestyle{empty}
\pagestyle{empty}

\begin{abstract}
Tactile Walking Surface Indicators (TWSIs) are safety-critical landmarks that blind and low-vision (BLV) pedestrians use to locate crossings and hazard zones. From our observation sessions with BLV guide dog handlers, trainers, and an O\&M specialist, we confirmed the critical importance of reliable and accurate TWSI segmentation for navigation assistance of BLV individuals. Achieving such reliability requires large-scale annotated data. However, TWSIs are severely underrepresented in existing urban perception datasets, and even existing dedicated paving datasets are limited: they lack robot-relevant viewpoints (e.g., egocentric or top-down) and are geographically biased toward East Asian \textit{directional bars}---raised parallel strips used for continuous guidance along sidewalks. This narrow focus overlooks \textit{truncated domes}---rows of round bumps used primarily in North America and Europe as detectable warnings at curbs, crossings, and platform edges. As a result, models trained only on bar-centric data struggle to generalize to dome-based warnings, leading to missed detections and false stops in safety-critical environments.
We introduce GuideTWSI, the largest and most diverse TWSI dataset, which combines a photorealistic synthetic dataset, carefully curated open-source tactile data, and quadruped real-world data collected and annotated by the authors. Notably, we developed an Unreal Engine–based synthetic data generation pipeline to obtain segmented, labeled data across diverse materials, lighting conditions, weather, and robot-relevant viewpoints. Extensive evaluations show that synthetic augmentation improves truncated dome segmentation across diverse state-of-the-art models, with gains of up to +29 mIoU points, and enhances cross-domain robustness. Moreover, real-robot experiments demonstrate accurate stoppings at truncated domes, with high repeatability and stop success rates (\textbf{96.15}\%). The GuideTWSI dataset, model weights, and code are publicly released in \href{...}{https://guidedogrobot-tactile.github.io/}. 

\end{abstract}

\section{INTRODUCTION}
\begin{figure}
    \centering
    \includegraphics[width=\linewidth]{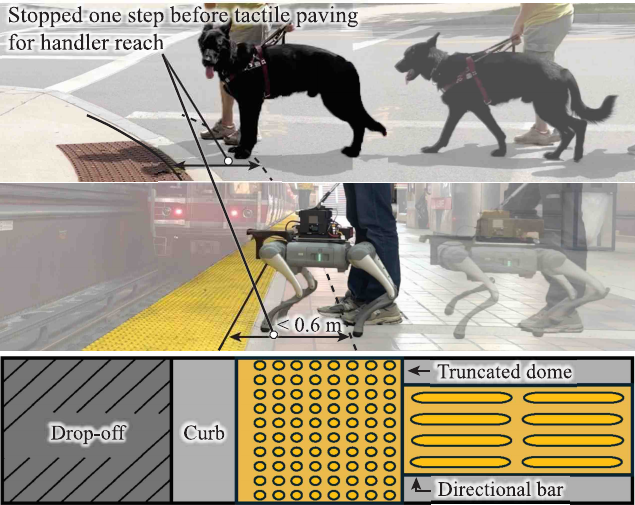}\
    \caption{\textbf{Objective of the proposed research and tactile walking surface indicator (TWSI).} (1) A guide dog stopping at a TWSI and curb, (2) a robot mimicking this stop behavior, and (3) different TWSI types.}
    \label{fig:starter}
\end{figure}

Tactile Walking Surface Indicators (TWSIs) provide vital environmental cues -- such as pedestrian crossing points, platform edges, hazard zones -- for blind and low-vision (BLV) pedestrians~\cite{emerson2021tactile}. TWSIs are typically composed of truncated domes and directional bars (see Fig.~\ref{fig:starter}), offer multisensory feedback through touch, sight, and sound, and are codified in accessibility regulations (e.g., ADA~\cite{ada_gov}, California Title 24~\cite{cbc2022_11B705_1_1_2}, AASHTO M 333-16~\cite{aashto_m333_16}) specifying dome sizes, spacing, and etc. In orientation and mobility (O\&M) practice, such landmarks anchor spatial awareness and guide safe, efficient movement toward a destination~\cite{giudice2008blind}. Despite their critical role, TWSIs remain severely underrepresented in urban perception datasets, limiting the ability of autonomous mobility assistive robots to reason about sidewalk semantics across diverse viewpoints, layouts, and lighting conditions.


Several datasets have been introduced to advance sidewalk perception, including SideGuide~\cite{park2020sideguide}, Tenji10K~\cite{takano2024tactile}, and TP~\cite{zhang2024grfb}. Takano et al.~\cite{takano2024tactile} released a dataset of 20 sequences comprising 10K first-person directional bar images taken in Japan for detection and tracking. Zhang et al.~\cite{zhang2024grfb} proposed a multi-scale feature extraction module for UNet-based segmentation~\cite{ronneberger2015u} and open-sourced a dataset of 1.4K samples captured under various appearances and lighting.

\begin{figure*}
    \centering
    \includegraphics[width=\textwidth]{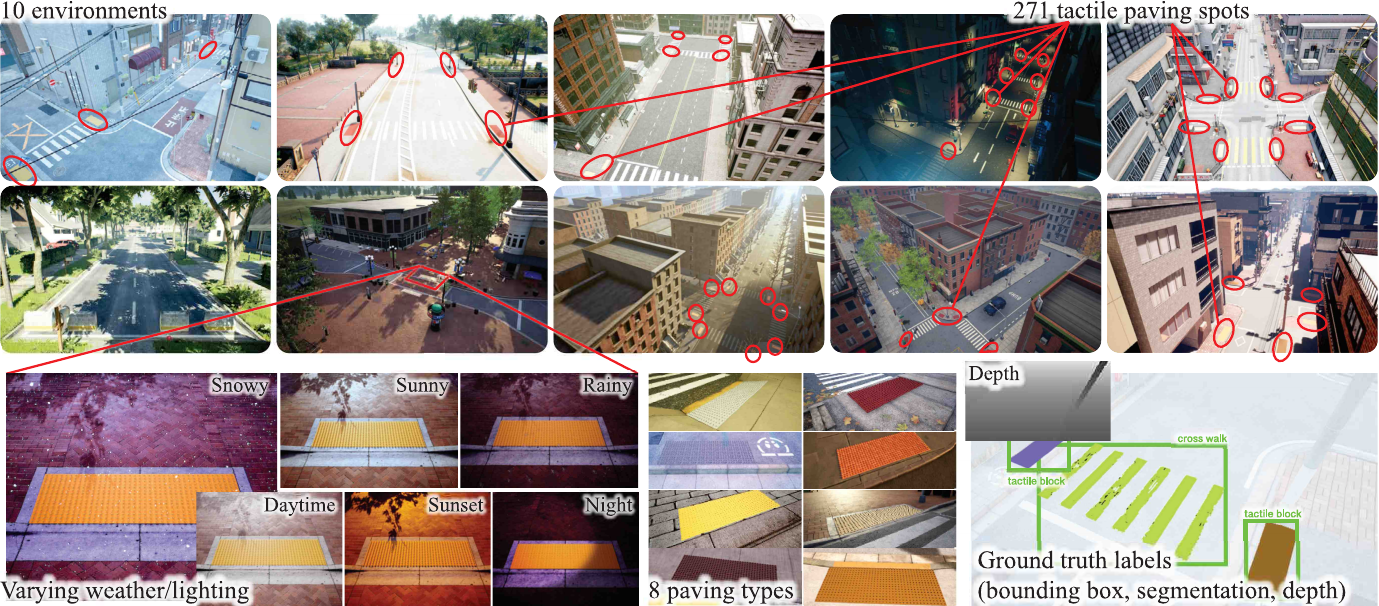}
    \caption{\textbf{Photorealistic synthetic tactile paving dataset generation.} Our pipeline uses various sidewalk environments in Unreal Engine 4 with varying viewpoints, lighting, and weather conditions to simulate real-world variability. We use AirSim for automatic annotation of depth, instance masks, and bounding boxes. This enables the creation of a high-quality dataset of 15K samples that significantly enhances TWSI segmentation when fused with real-world data.}
  \label{fig:main}
\end{figure*}
While valuable, these works are mostly constrained to first-person viewpoints, which are higher than those of ground robots like quadrupeds and cannot capture other perspectives---such as a top-down chin-mounted view---that are critically important for accuracy-critical tasks (e.g., stopping at truncated domes). Moreover, these works are geographically biased toward East Asia, where directional bars dominate for continuous path indicators. By contrast, in the United States and most of Europe, focus on truncated domes to be used at high-stakes decision zones such as curb ramps, crossings, and transit edges~\cite{emerson2021tactile}. Current datasets rarely cover truncated domes and lack variations in robot-relevant viewpoints and environmental factors (e.g., lighting, weather, materials). This limits the robustness and generalization of models when deployed on autonomous sidewalk platforms such as guide robots~\cite{takagi2025field,cai2024navigating,hwang2024towards} and delivery bots.

To overcome these limitations, we built a new dataset, named \emph{GuideTWSI}, an extensive dataset tailored for TWSI segmentation. The dataset is composed of three efforts: (1) collection and labeling of real-world quadruped data (RDome-2K), (2) meticulous curation of existing open-source datasets (RBar-22K), and (3) synthetic data generated by our Unreal Engine 4-based simulation environment (SDome-15K). For real-world data collection, we used a quadruped robot with an RGB camera and labeled the resulting robot-perspective dataset. Open-source data often contains mixed labels and inconsistencies that make it incompatible with fine-tuning pipelines. We gathered SideGuide~\cite{park2020sideguide}, Tenji10K~\cite{takano2024tactile}, TP~\cite{zhang2024grfb}, and community repositories, then post-processed them to ensure consistency across datasets and compatibility with our training pipeline.

Among the three components, the most notable effort is the development of synthetic data generation pipeline, which offers scalable and customizable data creation in contrast to expensive human-collected real-world datasets. Built in Unreal Engine 4 (UE4) with AirSim~\cite{airsim2017fsr} for automated annotation, our framework produces high-diversity data spanning surface colors, lighting conditions, weather, and robot-relevant viewpoints as depicted in Fig.~\ref{fig:main}. From this pipeline, we created a large-scale dataset of over 15K photorealistic images across ten distinct scenes, each with dense pixel-level annotations. 

We conducted experiments to validate the effect of synthetic augmentation, which showed consistent segmentation performance gains across state-of-the-art architectures. Beyond the offline evaluation on image data, we conducted real-world robot experiments to demonstrate the improved downstream task reliability: stop success rate at 96.15\%. To our knowledge, this is the first work to achieve reliable stopping at truncated domes on a real guide robot, demonstrated across diverse, previously unseen real-world environments. 

In sum, our contributions are threefold:
\begin{itemize}
\item A photorealistic synthetic data generation pipeline, specifically designed for blind navigation, that simulates multiple truncated dome textures under diverse viewpoints and lighting conditions, generating over 15K samples.
\item A large-scale GuideTWSI dataset comprising (i) meticulously curated open-source annotations, (ii) our synthetic data, and (iii) real-world robot-collected data, all released with code and pretrained model weights.
\item Extensive evaluations of our synthetic dataset using state-of-the-art segmentation models, including ablation studies showing robustness across domains, and real-world experiments using a guide dog robot for stopping at truncated domes with a success rate over 96\%.
\end{itemize}








\section{RELATED WORK}
\subsection{Mobility Assistive Robots for Outdoor Navigation}
For the last five decades, researchers have explored robotic mobility aids for BLV individuals~\cite{meldog, kulyukin2004robotic, guerreiro2019cabot,zhang2023follower,hwang2024towards}. Noticeably, recent advances in robot hardware and AI are now enabling practical use cases of guide robots to navigate complex outdoor environments~\cite{takagi2025field,cai2024navigating}. For example, Takagi et al.~\cite{takagi2025field} conducted large-scale field trials of an “AI Suitcase” navigation robot in a museum and outdoor areas, involving over 2,200 people (about 25\% of whom were BLV people). These trials revealed challenges encountered in outdoor navigation, such as maneuvering over curbs, tile edges, and manholes. Another commendable system is RDog~\cite{cai2024navigating} – a quadruped guide dog robot that integrates mapping, obstacle avoidance, and multimodal feedback (force and voice cues) to lead BLV users across varied terrain. In comparative user studies, the RDog system enabled faster and smoother travel with fewer collisions than either a white cane or a smart cane, while also lowering users’ cognitive load. While promising, existing systems still do not fully address the unique dangers in outdoor travel (e.g., open subway platform pits or busy roadway crossings). In particular, the ability to autonomously make safety-critical decisions---such as stopping at a height change (drop-off) or other hazards to protect the user---remains insufficiently addressed. This paper investigates this gap by developing reliable stop functionality at TWSIs.

\subsection{Recognizing Tactile Walking Surface Indicators}
Cane users typically detect curbs by tactile feedback---either the cane tip drops off the edge, or it encounters TWSIs, such as truncated domes, just before the curb. Likewise, guide dogs are trained to halt at curbs or any significant drop in height to prevent falls~\cite{tucker1984}. A robotic guide should similarly recognize upcoming height changes or hazards and stop the user safely in time. In many countries, standardized TWSIs are installed at the onset of street crossings and platform edges as a clear warning of an upcoming level change. Accurately detecting these visual markers would allow a robot to pinpoint where to stop before a curb. Indeed, researchers utilized computer vision methods for automatic real-time detection of TWSIs to assist BLV travelers~\cite{takano2024tactile,zhang2024grfb}. For example, Takano et al.~\cite{takano2024tactile} collected Tenji10K, a first-person dataset of 10,000 images of Japanese tactile paving (``Tenji blocks''), and demonstrated detection and tracking methods on it. Hwang et al.~\cite{hwang2024synthetic} introduced a synthetic Tactile-on-Paving dataset designed to train object detectors for TWSI detection at a relatively small scale. Leong and Lim released SurDis~\cite{leong2022surdis}, a dataset of about 17K depth/stereo images capturing sidewalk discontinuities (curbs, steps, gaps, etc.) with multi-class annotations. Despite these datasets and detection methods, to our knowledge, no existing guide-robot system yet uses visual TWSI recognition to trigger a stop at curb or truncated domes. In other words, integrating reliable tactile indicator detection into a mobility aid (analogous to a trained guide dog) remains an open challenge for safe robotic guidance.

\begin{table}
  \centering
  \caption{Observation Session Participant Demographics}
  \label{tab:exploratory}
  \begin{tabular}{ccccc}
    \toprule
    \textbf{ID} & \textbf{Age} & \textbf{Gender} & \textbf{Vision Level}  & \textbf{\makecell[c]{Experience$^{*}$}}\\
    \midrule
    GH01 & 63  & F & Totally blind & 36 \\
    GH02 & 66  & F & Legally blind  & 9 \\
    \cmidrule(lr){1-5}
    GT01 & - & F &  & -\\ 
    GT02 & 54 & M & & 21\\ 
    GT03 & 35 & F &  & 4\\ 
    GT04 & 28 & F &  & 7\\ 
    \cmidrule(lr){1-5}
    OT01 & 55 & M & & 10 \\
    \bottomrule
\end{tabular}
\end{table}

\section{Lessons learned from observations} 


To ground our technical contributions in real-world needs, we conducted a formative study with BLV guide dog handlers, professional trainers, and an O\&M specialist, drawing on the lived experiences of BLV travelers and the expertise of trainers. Our aim was to understand common practices and challenges in outdoor navigation, how handler–dog teams collaborate for safe mobility, and where guide dog robots must provide equivalent or more reliable support. The study combined semi-structured interviews with live observation sessions under trainer supervision, yielding insights into how guide dogs are trained to stop and signal at critical decision points and how O\&M specialists guide BLV travelers.

\subsection{Formative Study Design}
\subsubsection{Participants}
We recruited two guide dog handlers (GHs; all with visual acuity of 20/200 or worse and at least six months of guide dog experience), four professional guide dog trainers (GTs), and one O\&M specialist (OT). Luckily, we were able to observe the process of a new guide dog matching with a handler who had over 36 years of experience, giving us a unique opportunity to study how an early-stage guide dog works in new environments. All trainers and the O\&M specialist had more than four years of professional experience. Demographic details are summarized in Table~\ref{tab:exploratory}.

\subsubsection{Procedure}

Observation sessions captured early-phase training of a handler–dog team under the supervision of a trainer or O\&M specialist, as well as outdoor traveling with experienced handlers. These sessions included stopping at truncated domes, curbs, and cross streets. Semi-structured interviews explored handler–guide dog interactions and stopping behaviors, complementing live observations.


\subsection{Findings}
\subsubsection{Where is the ideal stopping position?}  
Guide dog trainers consistently emphasized that guide dogs are conditioned to halt at ramped curbs, with truncated domes serving as the most reliable landmark: as one trainer (GT02) explained, ``this is an indicator for dogs, like a landmark.’’ While there are no formal guidelines on the exact distance, both trainers and the O\&M specialist agreed that the safest position is approximately one stride back from the curb — far enough to avoid the drop-off, yet close enough for the handler to confirm the edge within a single step forward.

\subsubsection{Challenges in reliable stopping}  
Our observations highlighted the variability of stop accuracy across training stages. GH02, an experienced handler with a mature guide dog (six years), demonstrated near-perfect stopping behavior at curbs and truncated domes. In contrast, GH01---a more experienced handler paired with a newly trained dog—faced frequent inconsistencies: across 14 trials, the dog overstepped five times (passing the domes) and stopped too early four times (out of reach with one foot). These findings underscore that even with extensive handler expertise, early-stage guide dogs can misjudge tactile cues, leaving users vulnerable. This gap highlights the need for a safe decision-making system that can reliably stop at safety-critical locations.

\begin{table}[t]
  \centering
      \caption{GuideTWSI dataset comparison.} 
  \label{tab:dataset_comparison}
  \resizebox{\linewidth}{!}{
  \begin{tabular}{lcccc}
    \toprule
    \textbf{Dataset} & \textbf{Scale} & \textbf{Type} & \textbf{Geography} & \textbf{Modalities} \\
    \midrule
    SideGuide~\cite{park2020sideguide} & $\sim$8.2K & Real/bars & Korea & RGB, BBX, Seg. \\
    Tenji10K~\cite{takano2024tactile} & 10K & Real/bars & Japan & RGB, Seg. \\
    TP~\cite{zhang2024grfb} & $\sim$1.4K & Real/bars & China & RGB, Seg. \\
    \midrule
    \textbf{RBar-22K (compiled)} & \textbf{$\sim$22K} & \textbf{Real/bars} & Mostly Asia & \textbf{RGB, Seg.} \\
        \midrule
    \textbf{RDome-2K (ours)} & \textbf{2.4K+} & \textbf{Real/domes} & United States & \textbf{RGB, Seg.} \\
    \textbf{SDome-15K (ours)} & \textbf{15K+} & \textbf{Synthetic/domes} & Simulated & \textbf{RGB+D, BBx, Seg.} \\
    \bottomrule
  \end{tabular}}
\end{table}

\section{Synthetic Truncated Dome Data Generation}

We present a photorealistic synthetic data generation pipeline tailored for sidewalk navigation and accessibility perception. Our pipeline is built in Unreal Engine 4 (UE4) and leverages Microsoft’s AirSim~\cite{airsim2017fsr} for automated ground truth annotation. Our approach enables the creation of diverse task-specific synthetic scenes with rich labels. Specifically, we vary environmental conditions (e.g., lighting, weather, and textures) to mimic real-world variability. For example, we randomly change sun position, brightness, fog/rain effects, and material properties in each scene. Likewise, we apply texture and color randomization on surfaces to narrow the sim-to-real gap. In summary, our customizable pipeline can generate synthetic images with annotations under many conditions to train robust perception models.

\subsection{Environment and objects}
We base our scenes on ten UE4 environment assets available from Fab~\cite{fab}, to cover diverse real-world scenarios. For instance, the \textit{City Park} environment includes mixed terrain (e.g., grass, gravel, pavement) with trees and paths, while the \textit{Downtown West} environment includes streets, curbs, vehicles, traffic lights, and benches. Each environment is populated with obstacles and context objects (e.g., cars, street objects) to reflect realistic sidewalks. We render each scene under multiple lighting/weather conditions (e.g., sunny midday, overcast, dusk, light rain) by adjusting the properties of UE4’s Directional Lights and effects, randomly sampling parameters for each rendering batch. 

In addition to the base environments, we create custom truncated dome modules based on the Americans with Disabilities Act (ADA)~\cite{ada_gov} dimensions, and import them into each scene. The modules are textured with high-resolution materials (from a third-party \textit{Tactile Blocks} pack) so that they look photorealistic. To capture the diversity of real truncated domes, we vary the color of each tactile block (e.g., standard yellow, red, white, or gray variants). This lets our synthetic data include the broad range of appearances of truncated domes. Note that our pipeline is not limited to TWSIs; similarly, additional accessibility cues (e.g., pedestrian audible signal buttons, accessible door buttons) can be incorporated by placing modeled objects with appropriate textures.


\subsection{Camera viewpoints}
Using UE4 and AirSim, we simulate multiple camera perspectives corresponding to different camera setups of mobile robots. For each environment, we program several trajectories that sweep the scene. In particular, we use three types of camera paths: (1) a circular path that orbits the target object (e.g., truncated domes), and (2) a top-down sweep that moves a camera from an elevated position downward over the scene, more suitable for systems with bottom-facing cameras. These trajectories produce views from low to overhead angles, reflecting the variety of possible robot-mounted cameras. Along each path, we also vary the camera’s height and orientation to further increase viewpoint diversity. In total, these varied trajectories yield a broad spectrum of perspectives on each truncated dome installation, improving the generality of the trained models.

\subsection{Ground truth generation}
Our pipeline can automatically generate dense ground truth for every rendered frame. We assign a unique semantic label ID to each object (including tactile domes and all other environment elements) in UE4. AirSim uses these IDs to produce pixel-level semantic and instance segmentation masks for every object. At the same time, AirSim drives the virtual cameras to capture synchronized RGB images and depth maps from the predefined trajectories. AirSim also outputs 2D bounding boxes for each labeled object in view. Using the bounding box and color-encoded mask information, we further extracted masks from the color channels associated with each semantic label ID and reformatted them into standardized labels tailored for various model training as detailed in Section~\ref{sec:realdata}. As a result, each synthetic image is paired with: (1) a semantic segmentation mask (class and instance labels for every pixel), (2) a list of 2D bounding boxes for all objects, (3) a depth map aligned to the RGB image, and (4) the camera’s intrinsic parameters. This rich annotation comes at minimal manual cost, since it is produced automatically by the simulation.

\subsection{Synthetic dataset specifications} 
Using our synthetic data generation pipeline, we created the \textbf{S}ynthetic truncated \textbf{Dome} (\textbf{SDome-15K}) dataset, with over 15,010 photorealistic images of truncated domes. Each sample includes RGB images, pixel-wise segmentation mask, depth, and 2D bounding boxes (COCO-style~\cite{lin2014microsoft}). Data span ten diverse environments with varied lighting, weather, and viewpoints relevant to robot-mounted cameras (circular and top-down). Custom truncated dome assets were modeled to comply with ADA standards and textured with realistic appearances, establishing SDome-15K as a large-scale benchmark for training TWSI perception models (see Table~\ref{tab:dataset_comparison}).

\section{REAL WORLD TACTILE INDICATOR DATA}
To evaluate the effectiveness of our synthetic dataset and to support real-world deployment, we (1) curated scattered existing open source tactile walking surface indicator data and (2) collected a robot-perspective data specified for truncated domes. Together, these datasets provide a large-scale, diverse benchmark for safety-critical TWSI segmentation.

\begin{figure*}
    \centering
    \includegraphics[width=\linewidth]{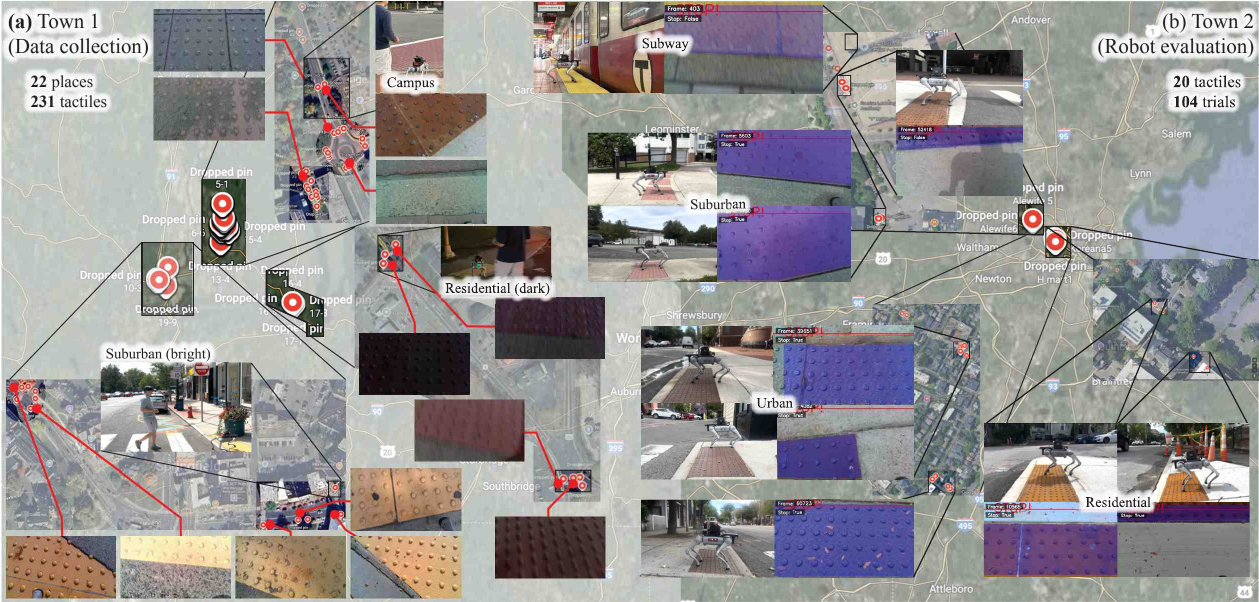}
    \caption{\textbf{Real robot data collection and hardware experiment.} 
    (a) We collected data with a quadruped robot across multiple sites in suburban, campus, and rural environments featuring various truncated domes. Data were gathered at different times of day under diverse truncated dome appearances. (b) We evaluated the robot's reliable stopping at truncated domes in unseen areas. We present the hardware experiment setup and segmentation inference results.}
    \label{fig:realrobotmap}
\end{figure*}

\subsection{Curated real-world tactile indicator data}
\label{sec:realdata}
We compiled TWSI samples from multiple public sources: the SideGuide dataset~\cite{park2020sideguide}, Tenji10K~\cite{takano2024tactile}, TP~\cite{zhang2024grfb}, and 69 community-contributed image sets on Roboflow~\cite{roboflow}. We filtered and unified these heterogeneous data source to create a single large-scale TWSI dataset. Specifically, from SideGuide we extracted only images containing the tactile paving class. From Tenji10K and TP, we removed redundant or noisy annotations. And on Roboflow, we identified relevant images by searching terms such as ``truncated domes'', ``tactile paving'', and ``braille block''.

Integrating these heterogeneous resources required substantial processing. After gathering the data, we removed duplicates across sources and standardized the annotation formats. We also performed rigorous quality control, discarding samples without proper segmentation masks (e.g., those with only bounding boxes or missing labels). We resolved further inconsistencies and unified formats for model training by converting all resources into run-length encoding (RLE) formats for SAM2.1 and polygon-only annotations (having only class ID and polygon coordinates without bounding boxes) for YOLOv11-seg models. After conversion, we manually overlaid masks on the corresponding RGB images to verify annotation quality. In total, 785 samples were excluded during curation, resulting in 19,925 high-quality, mask-annotated images---named \textbf{RBar-22K}---the comprehensive real-world TWSI dataset.

\subsection{Robot-collected truncated dome data}
While \textbf{RBar-22K} contains a large amount of labeled data, most of it was collected in Asia, with the majority consisting of directional bars. To support fine-tuned segmentation model evaluation on truncated domes, we collected a new robot-perspective truncated dome dataset, named \textbf{RDome-2K}. A Unitree Go2 robot equipped with an Intel RealSense D435 camera facing the ground with a $70^{\circ}$ angle (see Fig.~\ref{fig:hw}) was remotely controlled across diverse environments (e.g., campus, residential, suburban, and rural areas) at different times of day. This setup allowed us to capture egocentric, top-down views of truncated domes under varied conditions (light changes, occlusions, and surface damage, as shown in Fig.~\ref{fig:realrobotmap}. All collected 2,466 RGB frames were manually annotated using Roboflow’s auto-segmentation tool, followed by human verification. In contrast to existing TWSI datasets that mainly feature directional bars from human perspectives, our robot-collected truncated dome data (RDome-2K) provides top-down egocentric views of truncated domes tailored for robotic mobility assistance. This data enables extensive evaluation of truncated dome segmentation, particularly to quantify the benefit of our synthetic SDome-15K data, and establish a realistic benchmark for TWSI segmentation.


\begin{table*}[!t]
\renewcommand{\arraystretch}{1.2}
\caption{Impact of synthetic data augmentation on truncated dome segmentation (RBar-train (+SDome-15K) → RDome-2K)}
\label{fig:quant}

\centering
\begin{tabular}{lccccccccc}
\toprule
& \multicolumn{4}{c}{\textbf{Real Data Only}} & \multicolumn{4}{c}{\textbf{Real + Synthetic Data}} & \\
\cmidrule(lr){2-5} \cmidrule(lr){6-9}
\textbf{Method} & \textbf{Prec.} & \textbf{Rec.} & \textbf{mAP50-95} & \textbf{mIoU} & \textbf{Prec.} & \textbf{Rec.} & \textbf{mAP50-95} & \textbf{mIoU} & \textbf{$\Delta$ mIoU} \\
\midrule
YOLOv11-seg-N    & 0.7958 & 0.6924 & 0.5934 & 0.6161    & 0.8718 & 0.8084 & 0.7288 & \textbf{0.7308} & \textbf{+0.1147} \\
YOLOv11-seg-X    & 0.8838 & 0.8204 & 0.7362 & 0.7389    & 0.9102 & 0.8588 & 0.8188 & \textbf{0.7887} & \textbf{+0.0498} \\
Mask2Former     & 0.9458 & 0.5975 & 0.4798 & 0.5777    & 0.9611 & 0.8669 & 0.7829 & \textbf{0.8375} & \textbf{+0.2598} \\
SAM2.1+UNet         & 0.8680 & 0.5165 & 0.3475 & 0.4789    & 0.9704 & 0.7031 & 0.5627 & \textbf{0.6883} & \textbf{+0.2094} \\
DINOv3+RegCls   & 0.9027 & 0.7804 & 0.6176 & 0.7322 & 0.8667 & 0.8924 & 0.6933 & \textbf{0.7926} & \textbf{+0.0604 }\\
DINOv3+EoMT     & 0.8141 & 0.6237 & 0.4828 & 0.5804 & 0.9305 & 0.9197 & 0.8492 & \textbf{0.8756} & \textbf{+0.2952} \\
\bottomrule
\end{tabular}
\end{table*}

\graphicspath{{figure_pic/}}

\newlength{\cellw}
\setlength{\cellw}{0.15\textwidth}
\newcommand{\img}[1]{\includegraphics[width=\cellw,height=\cellw,keepaspectratio]{#1.jpg}}

\begin{figure*}[t]
\centering
\setlength{\tabcolsep}{2pt}
\renewcommand{\arraystretch}{0.95}

\begin{tikzpicture}
\node (table) {
\begin{adjustbox}{max width=\textwidth}
\begin{tabular}{c c c c c c c}
   & \multicolumn{3}{c}{\footnotesize\textbf{\vspace{0.15em}Real Data Only}}
   & \multicolumn{3}{c}{\footnotesize\textbf{\vspace{0.15em}Real + Synthetic Data }} \\
   \img{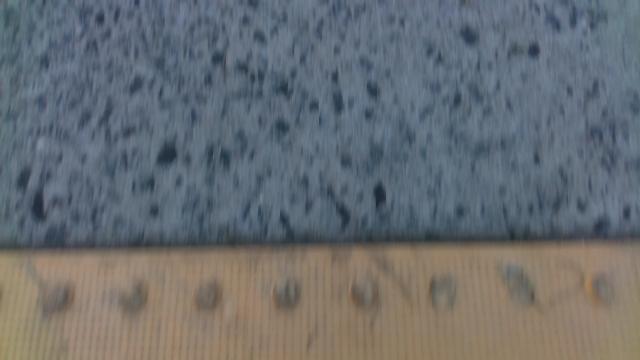} & \img{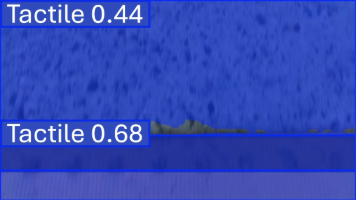} & \img{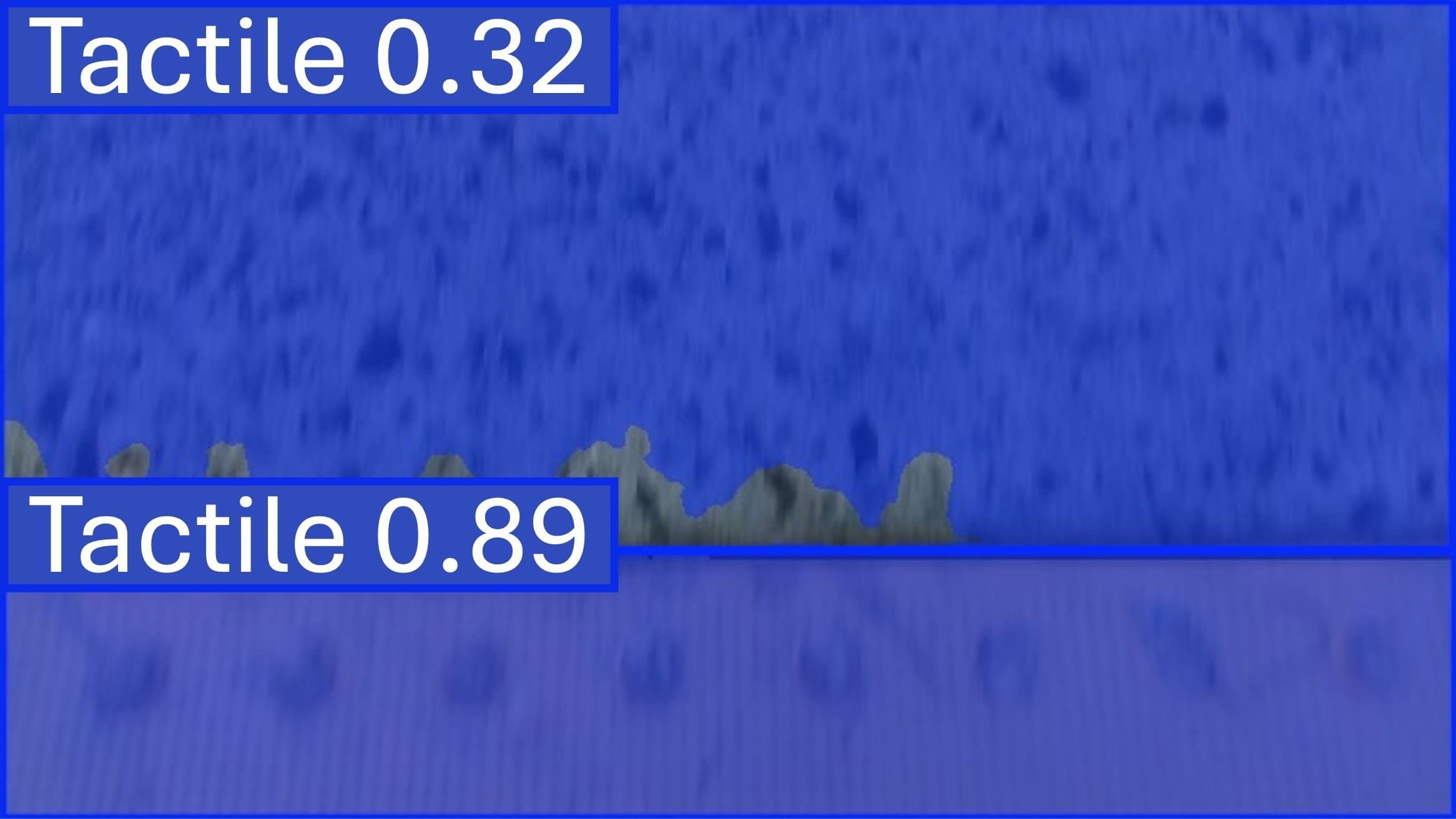} & \img{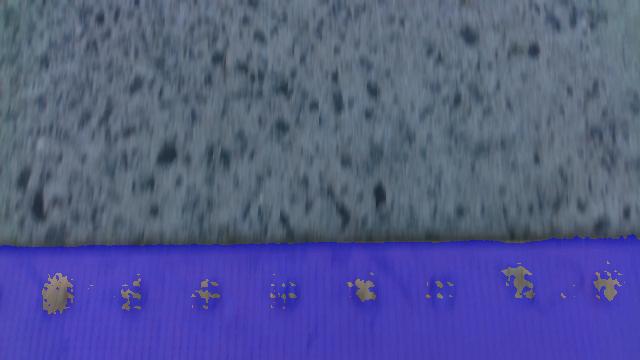}
           & \img{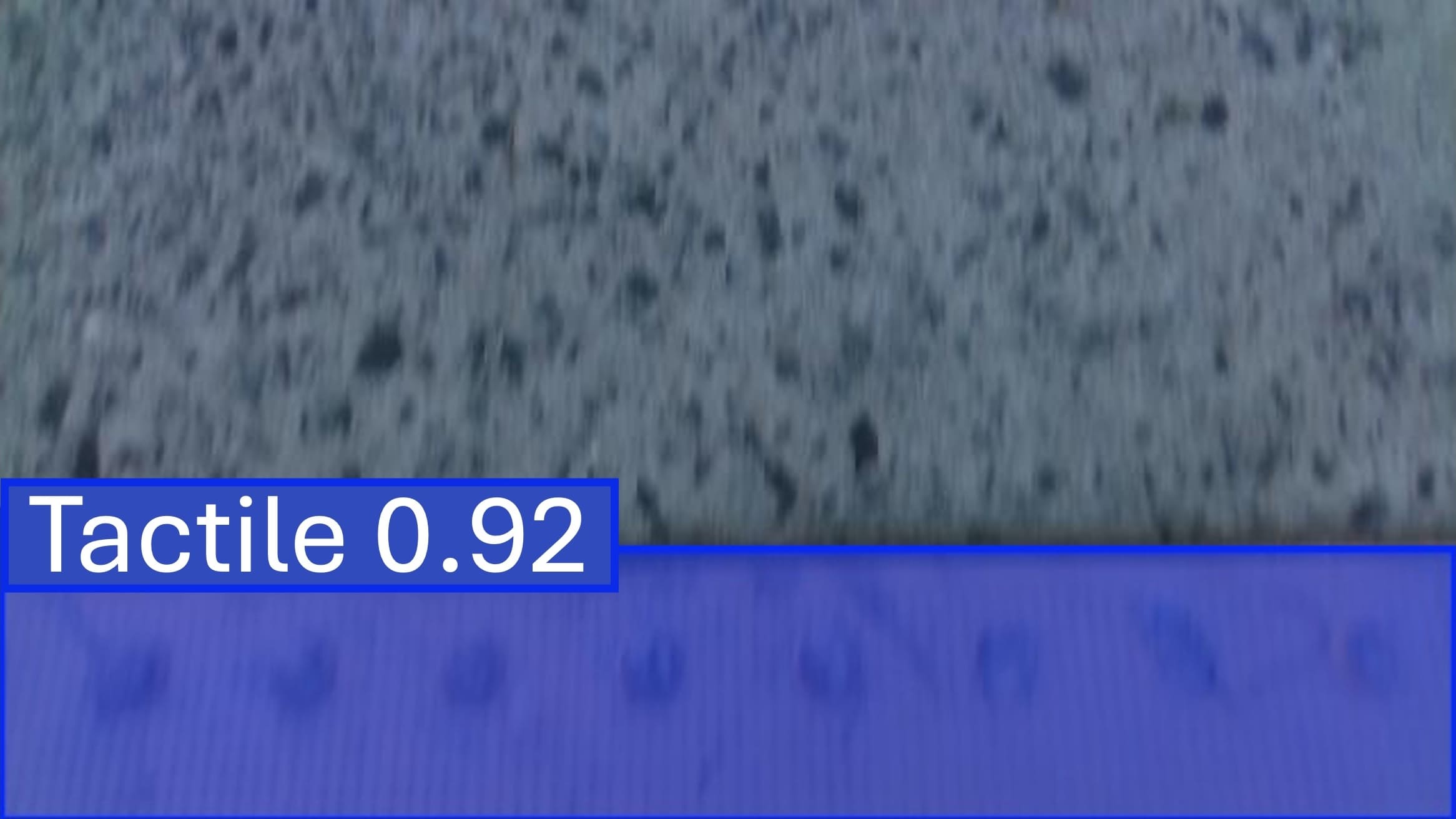} & \img{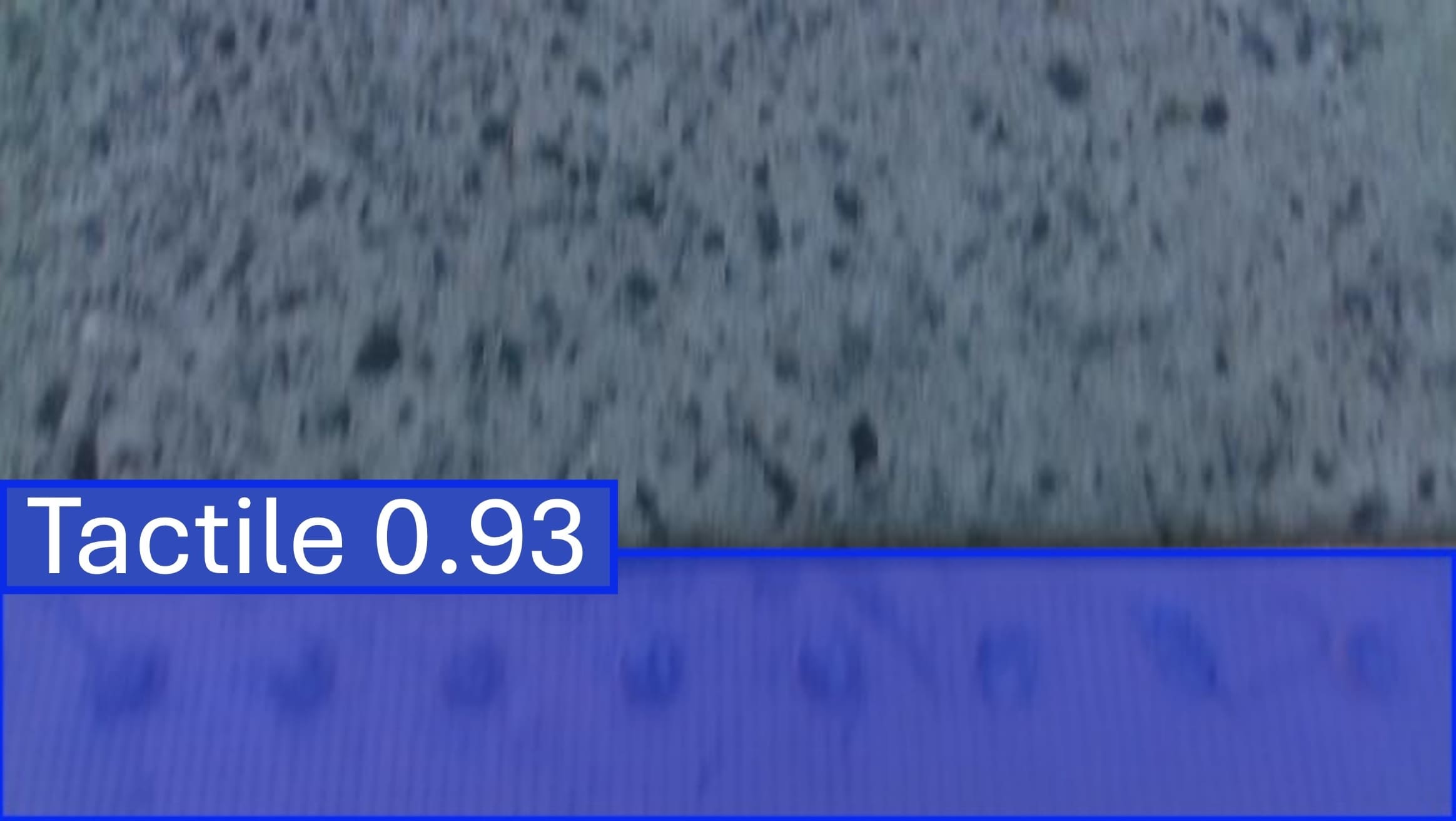} & \img{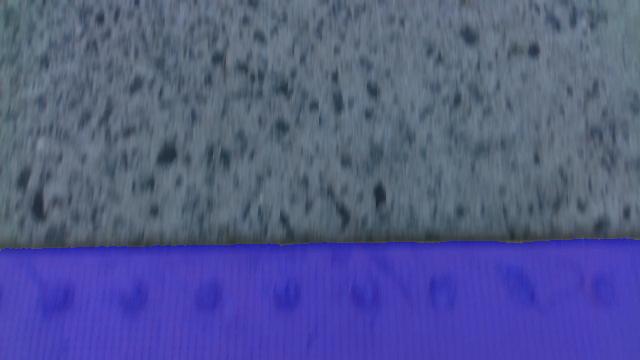} \\
   \img{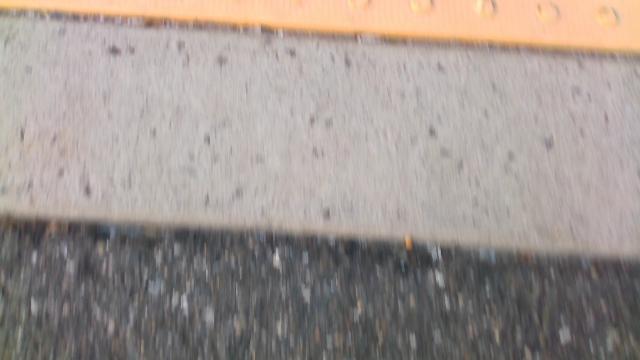} & \img{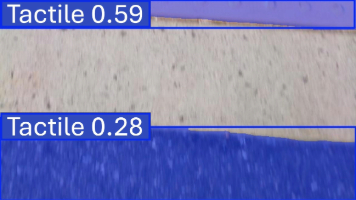} & \img{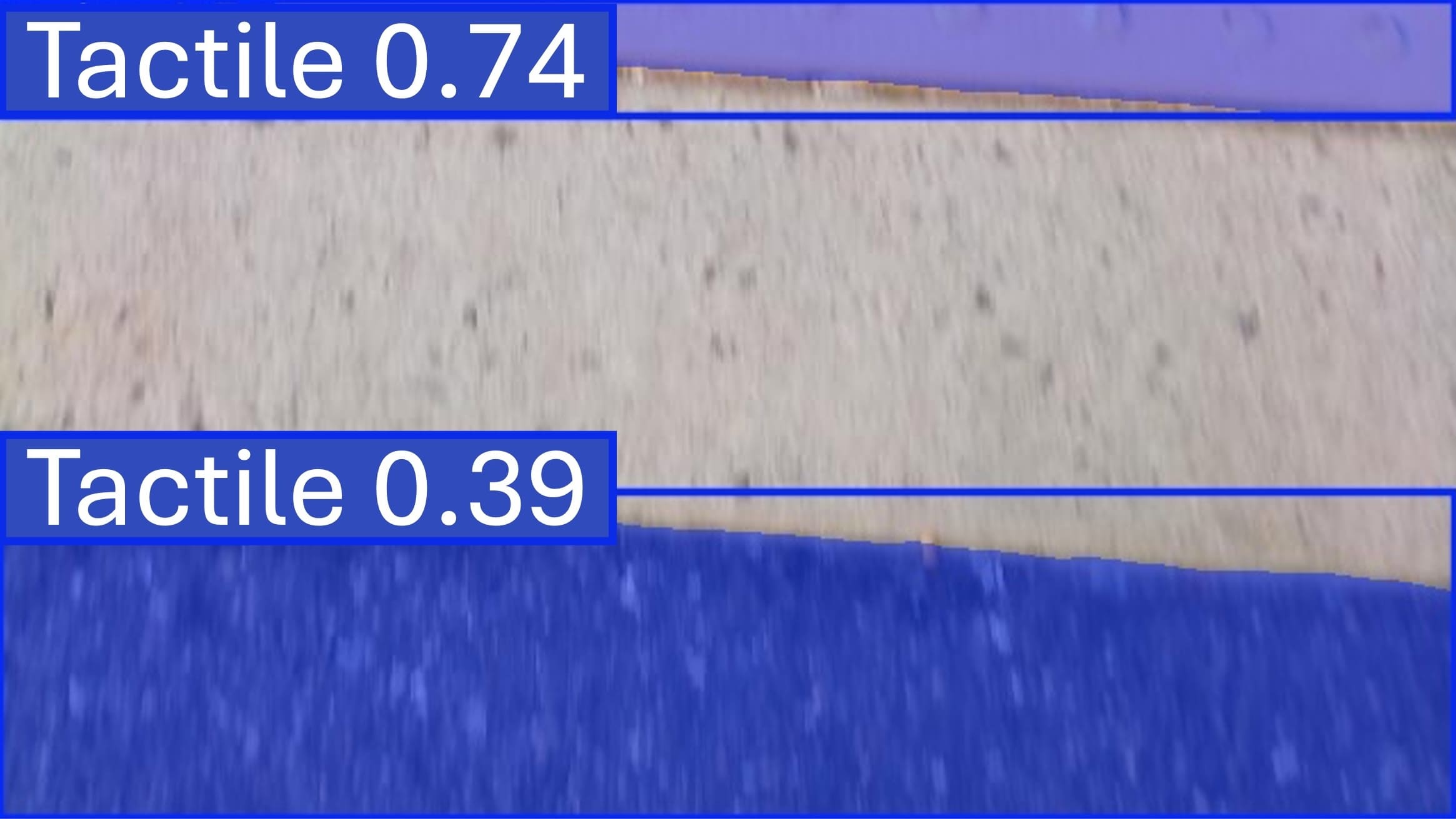} & \img{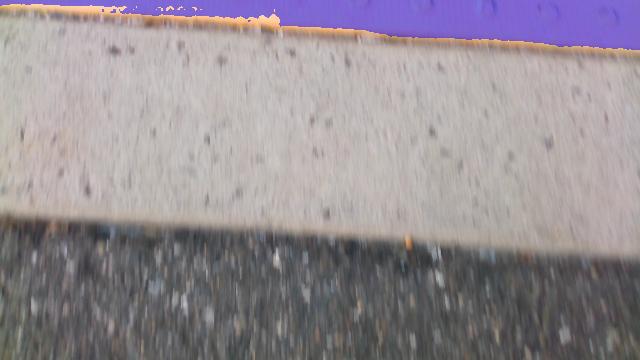}
           & \img{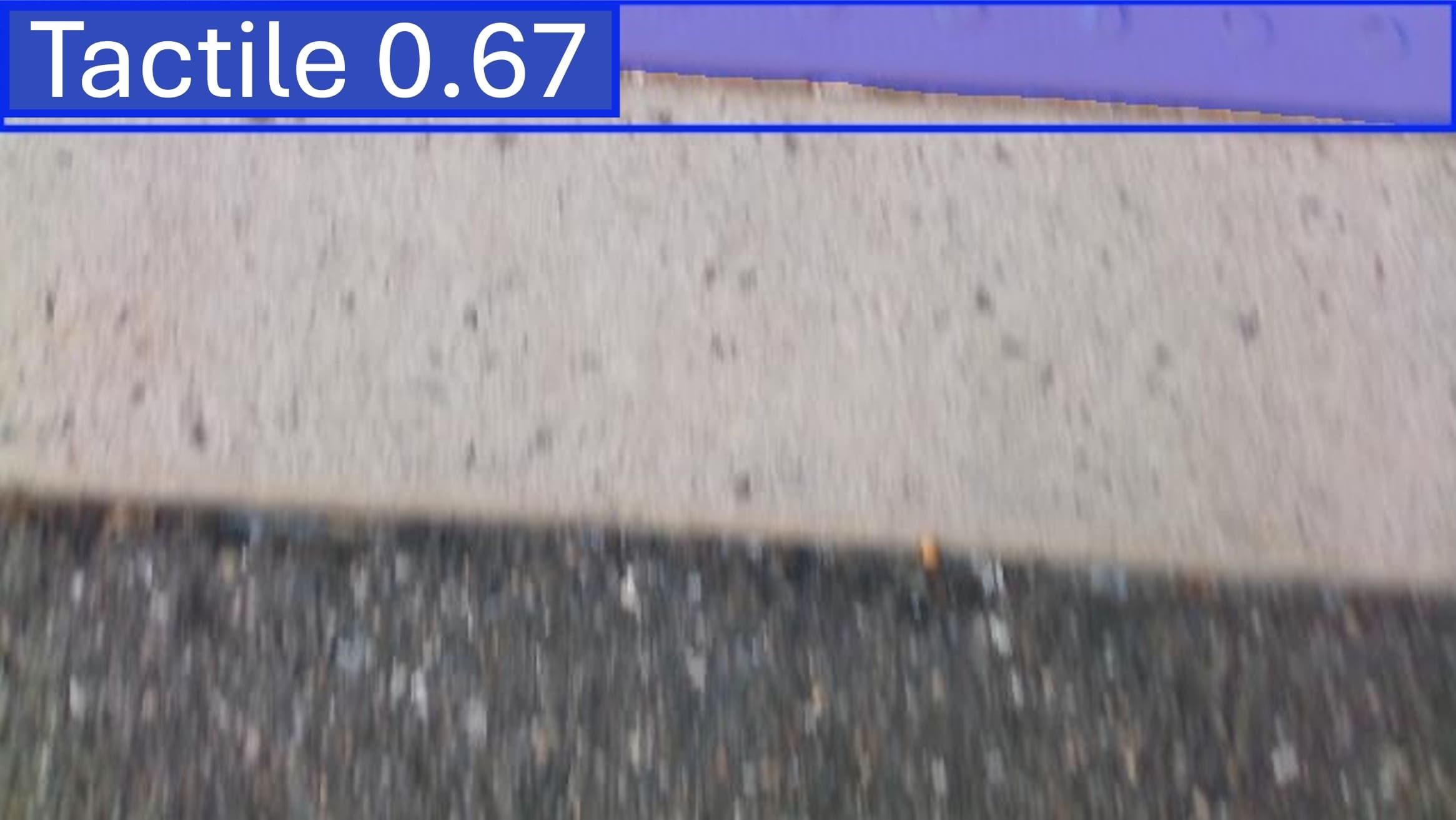} & \img{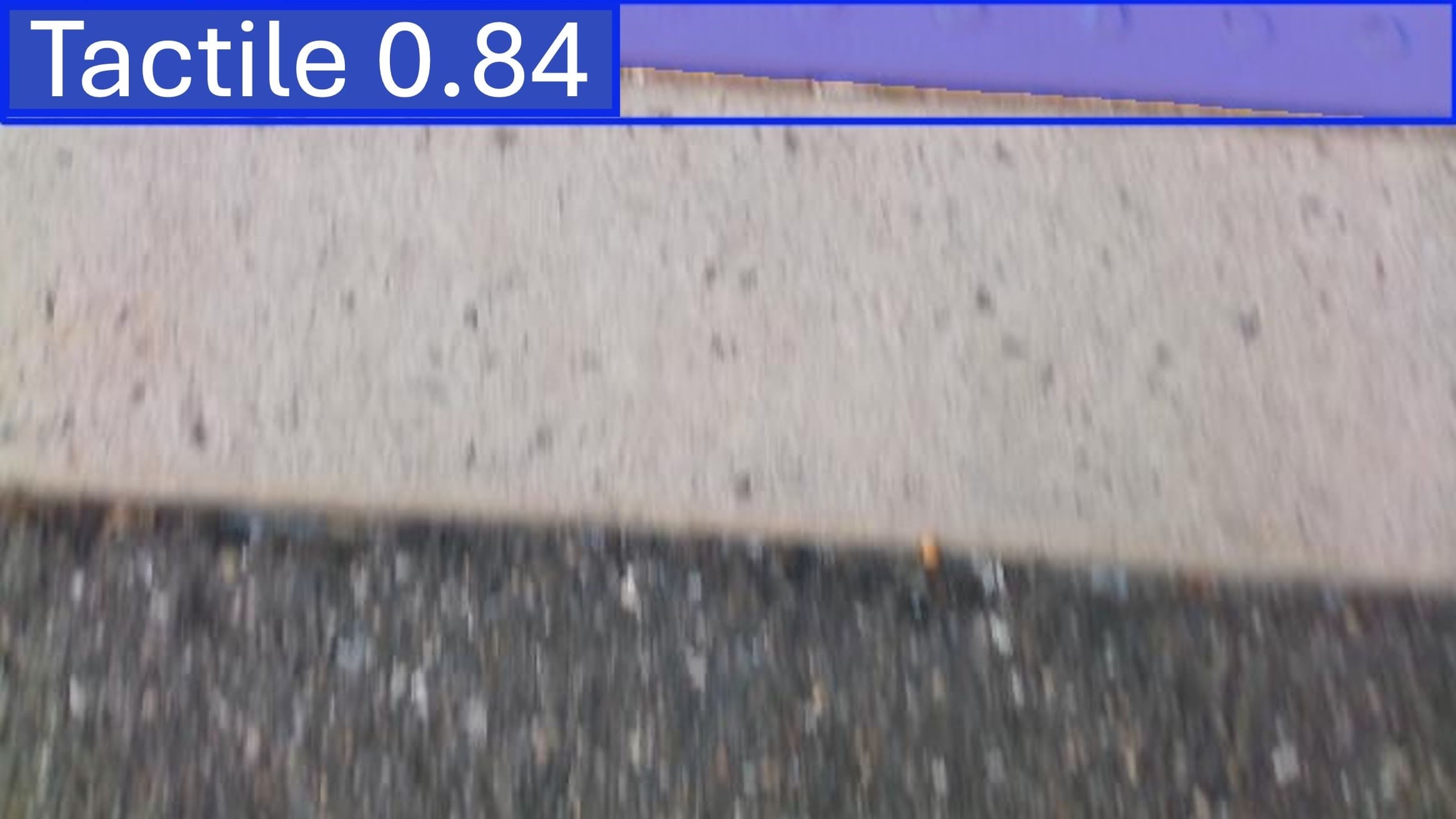} & \img{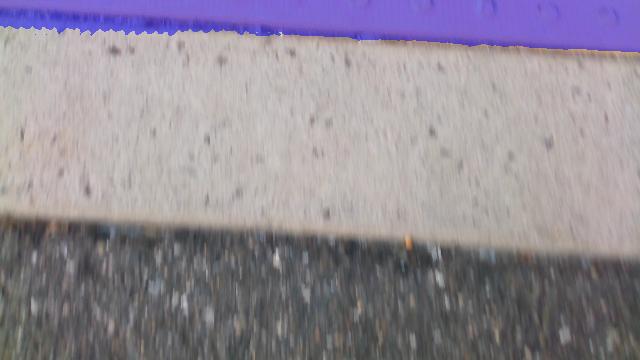} \\
   \img{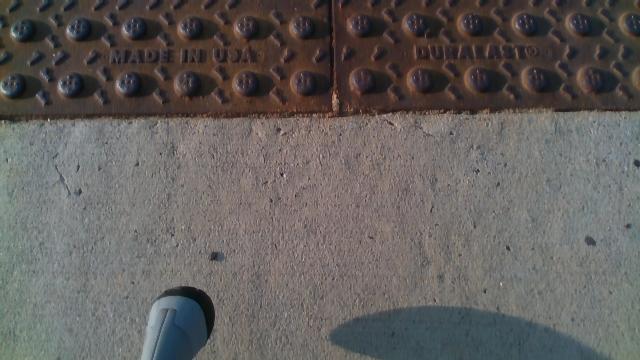} & \img{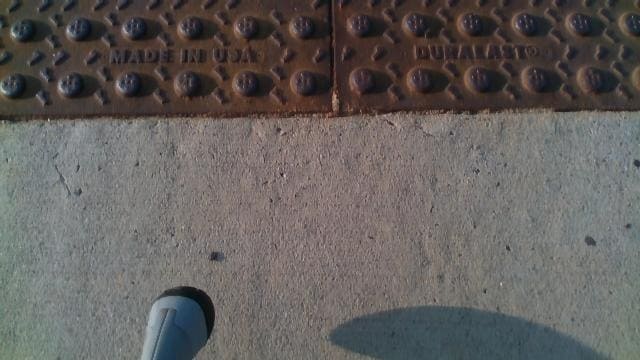} & \img{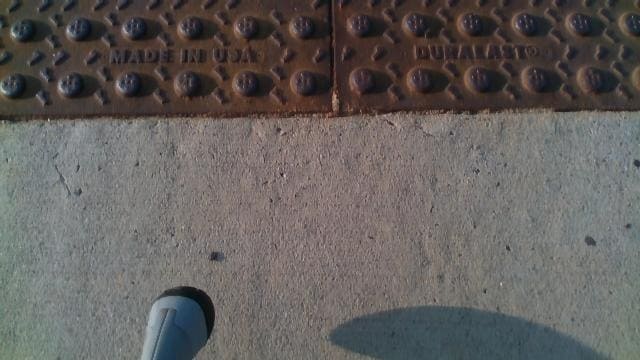} & \img{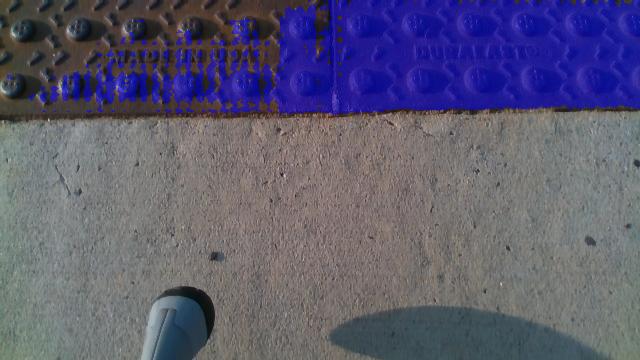}
           & \img{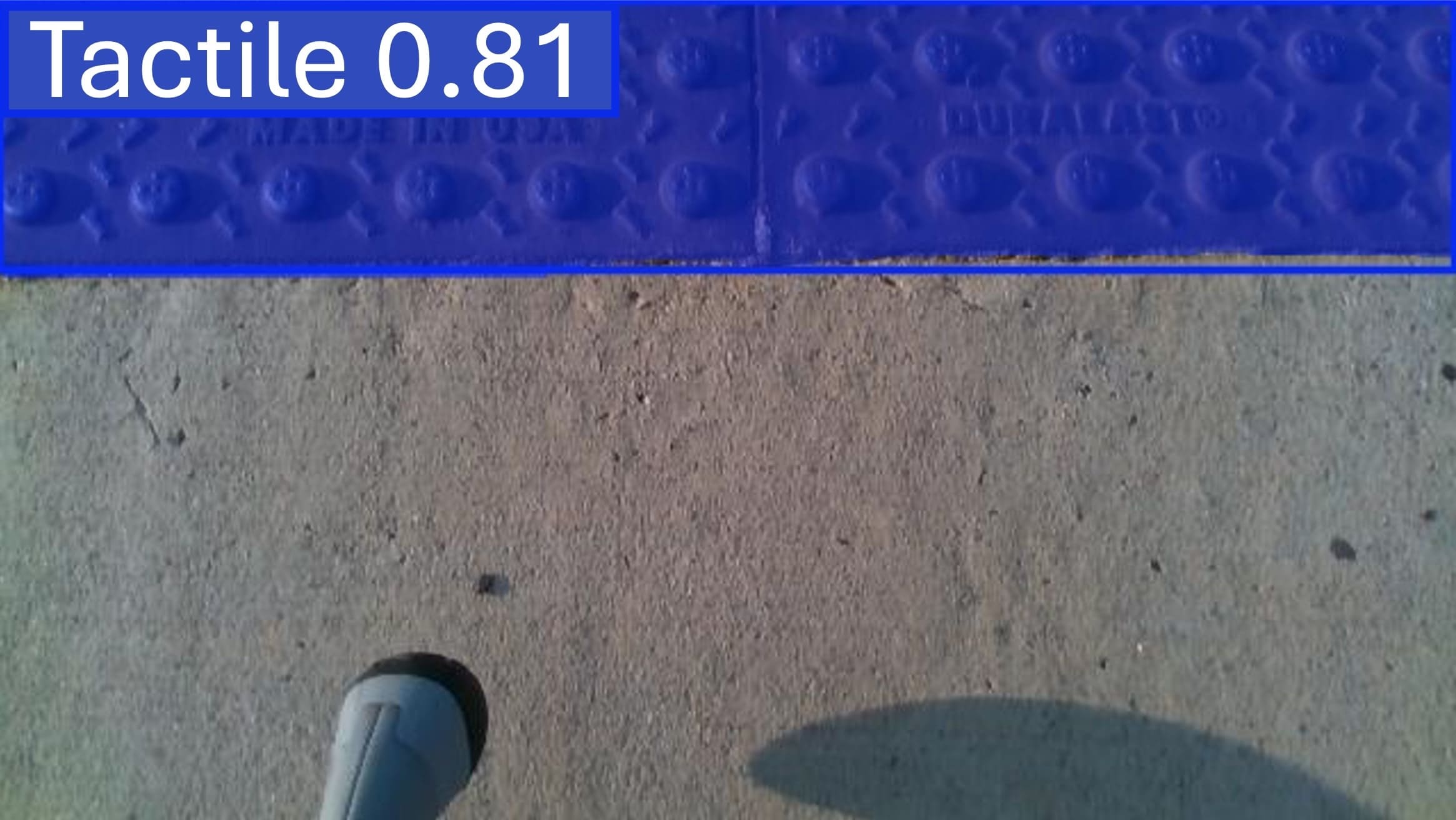} & \img{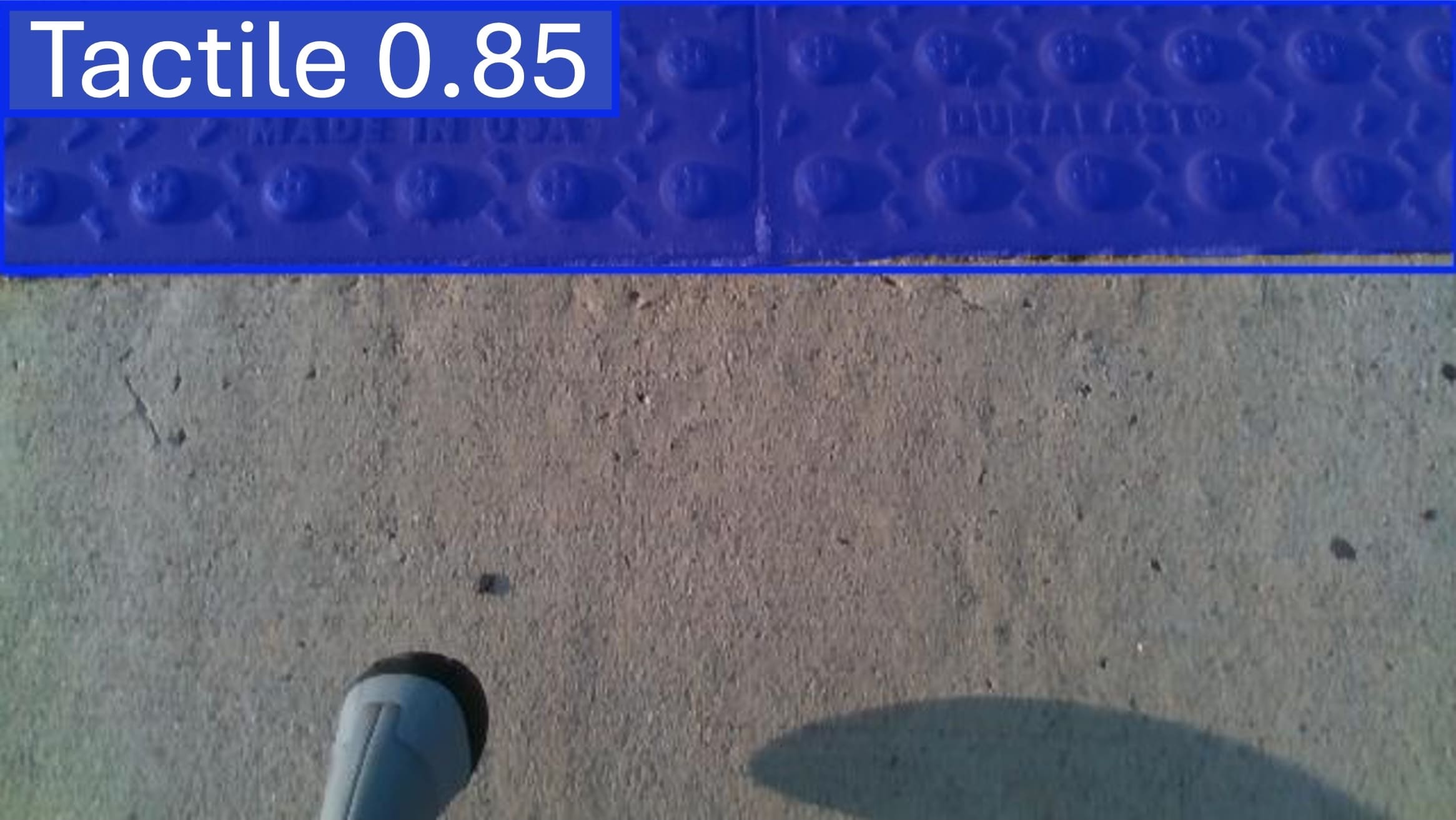} & \img{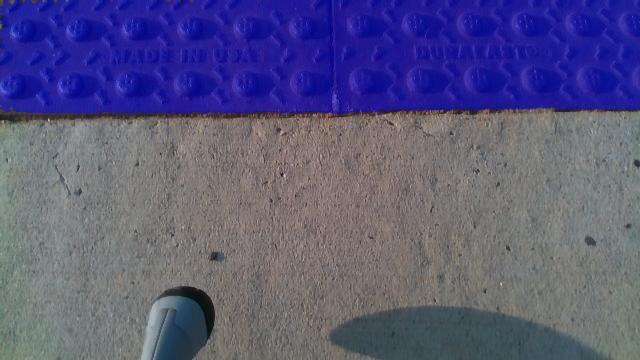} \\
   \img{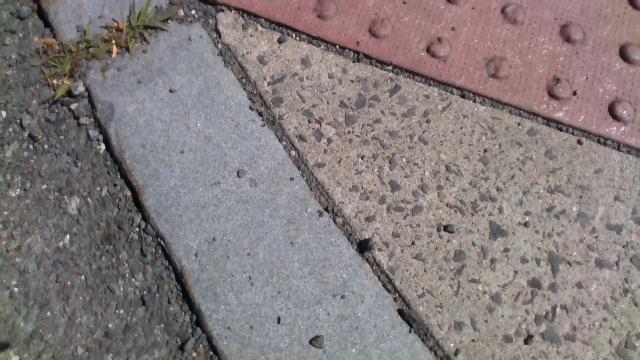} & \img{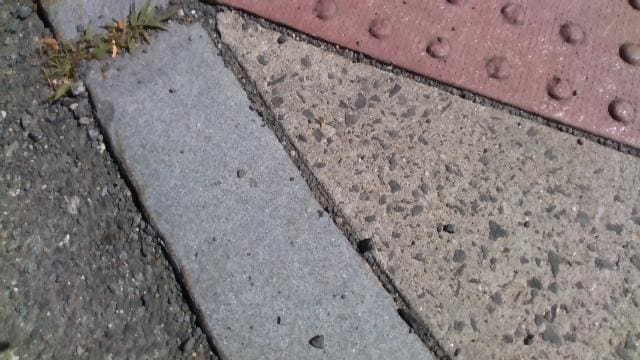} & \img{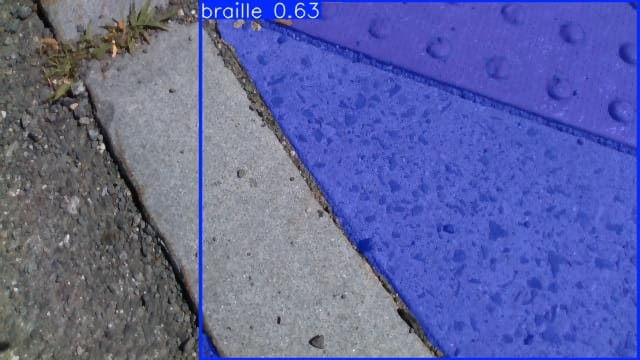} & \img{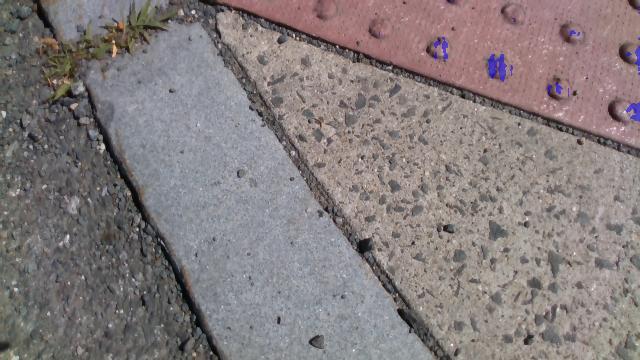}
           & \img{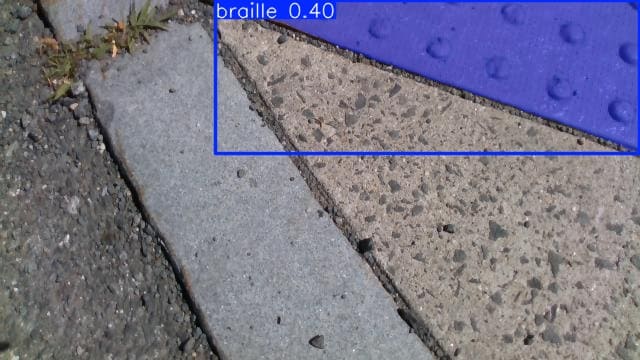} & \img{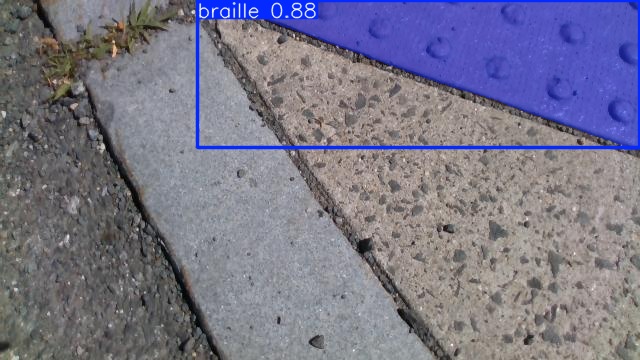} & \img{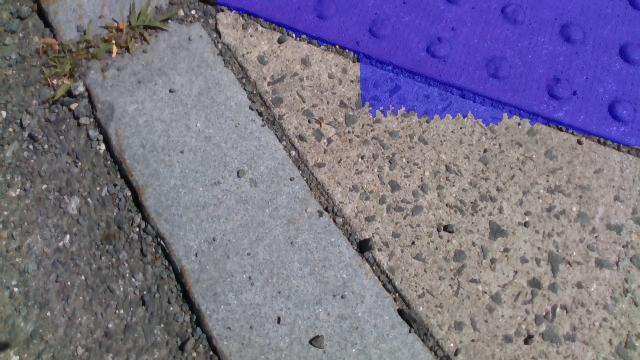} \\
   \img{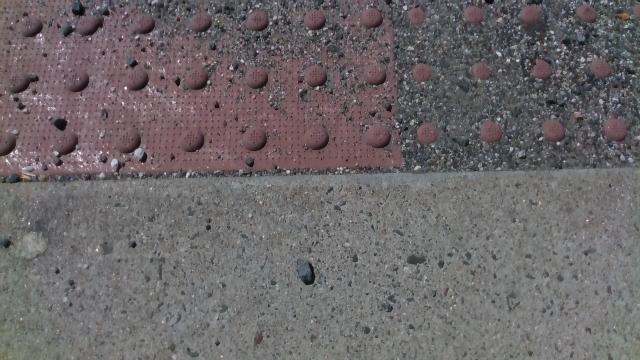} & \img{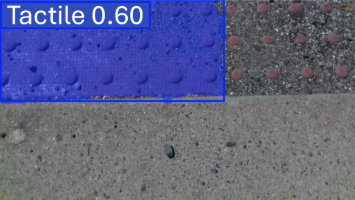} & \img{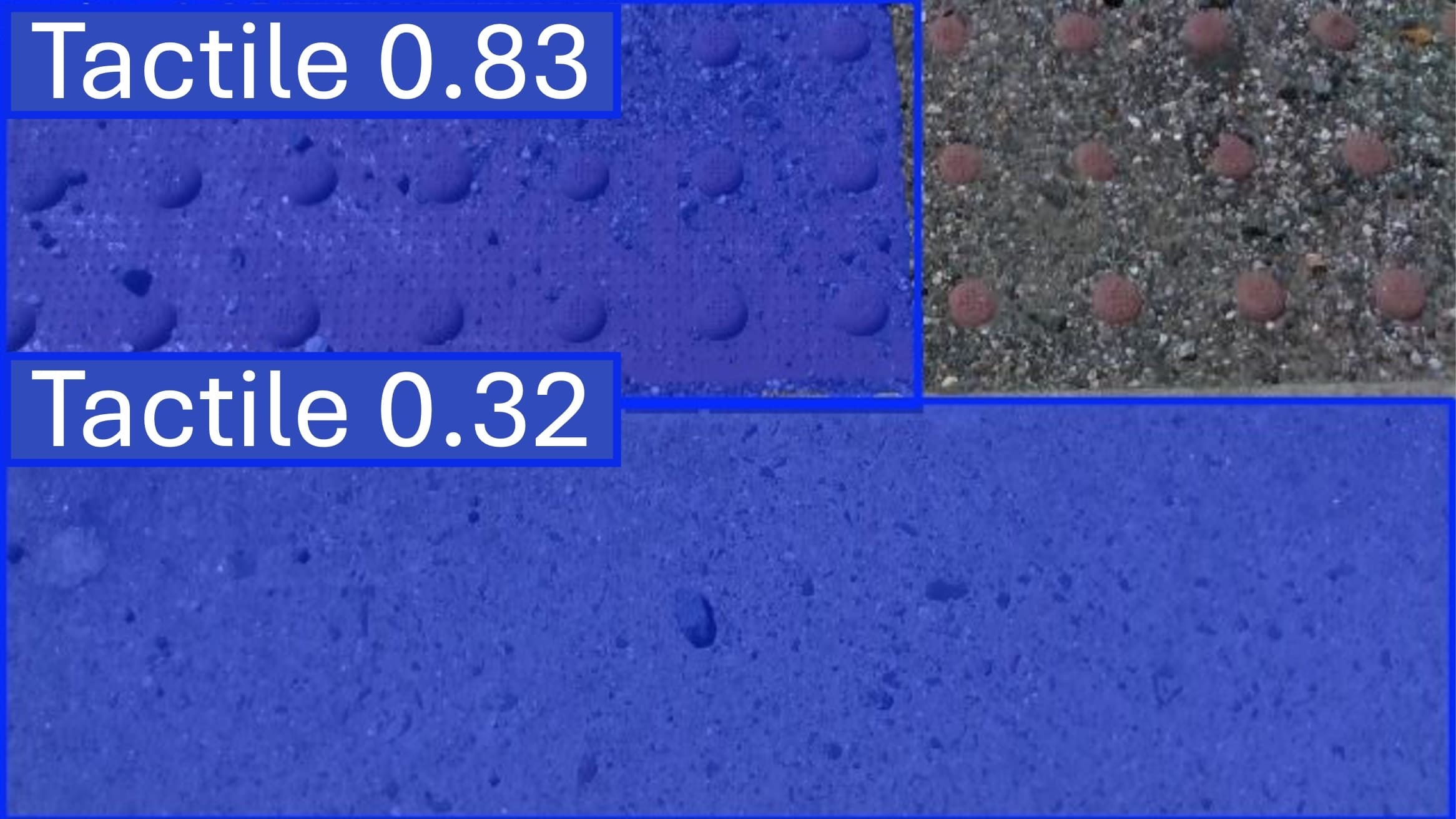} & \img{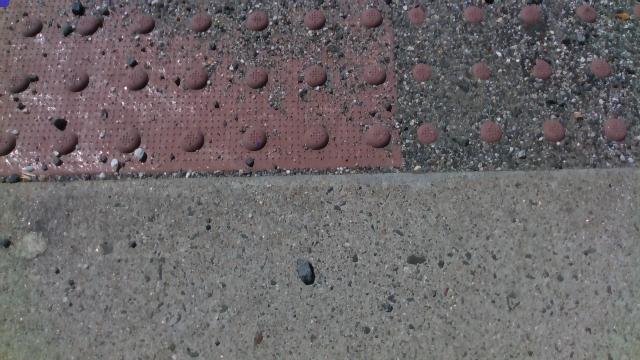}
           & \img{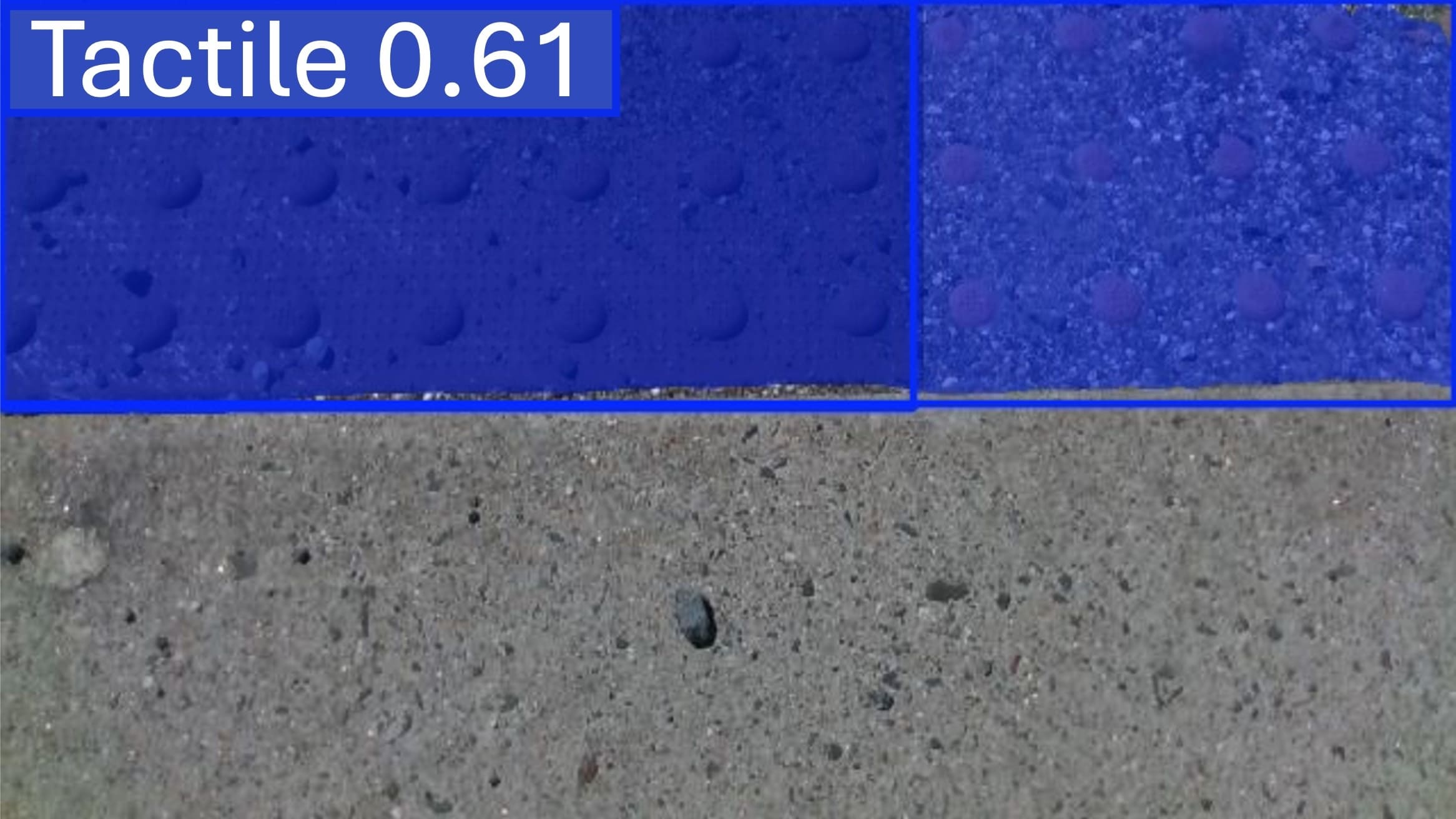} & \img{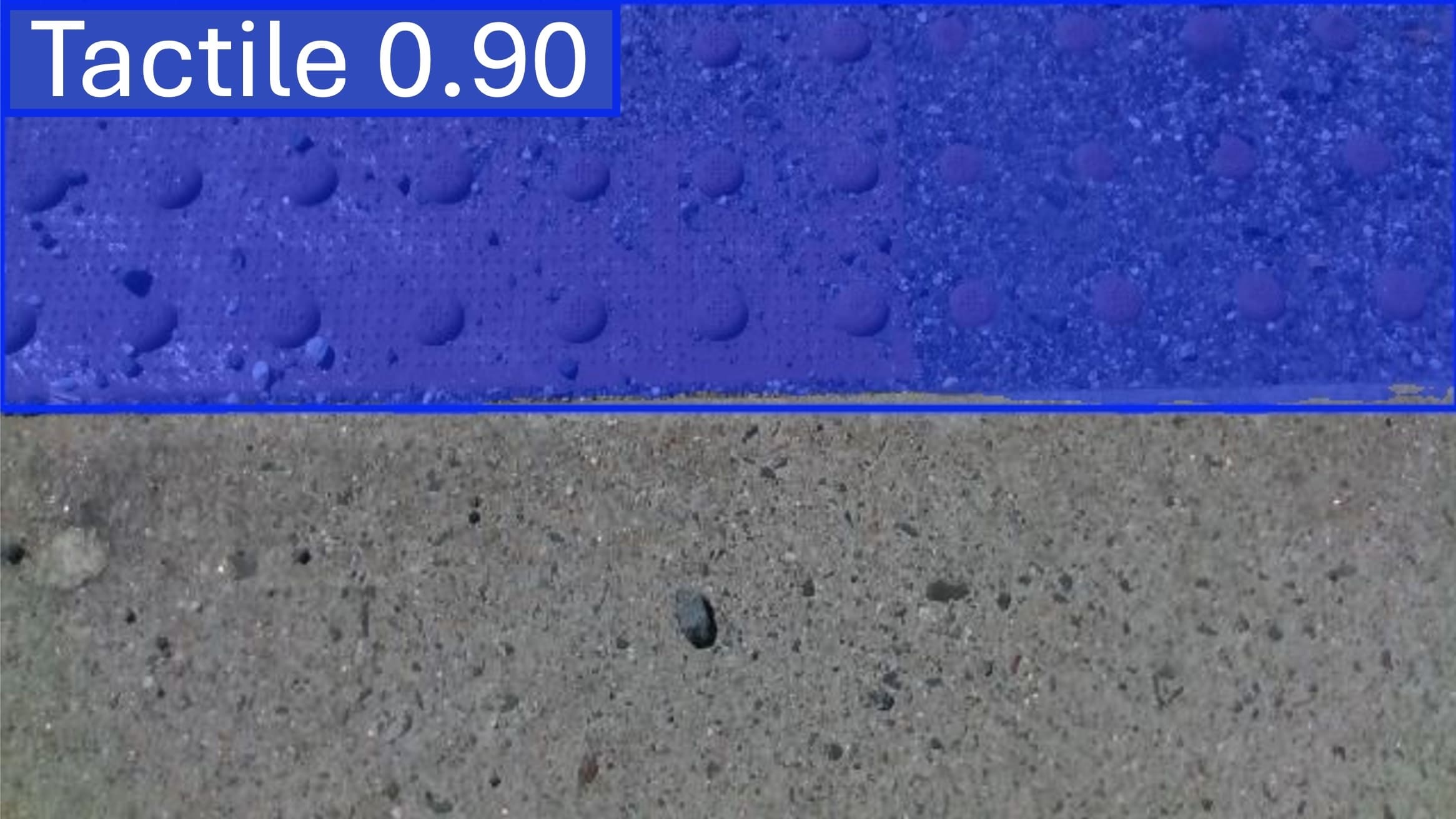} & \img{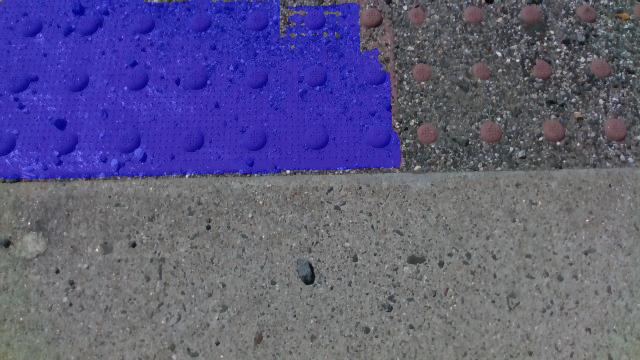} \\
   \img{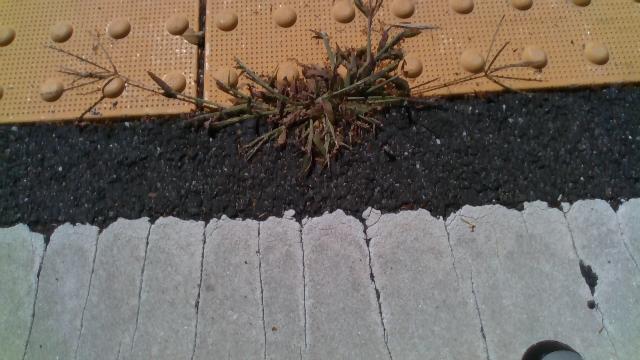} & \img{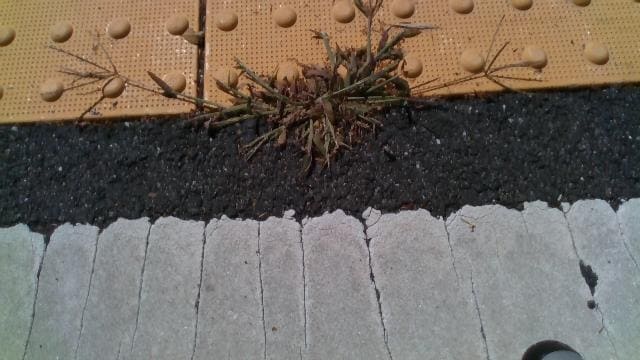} & \img{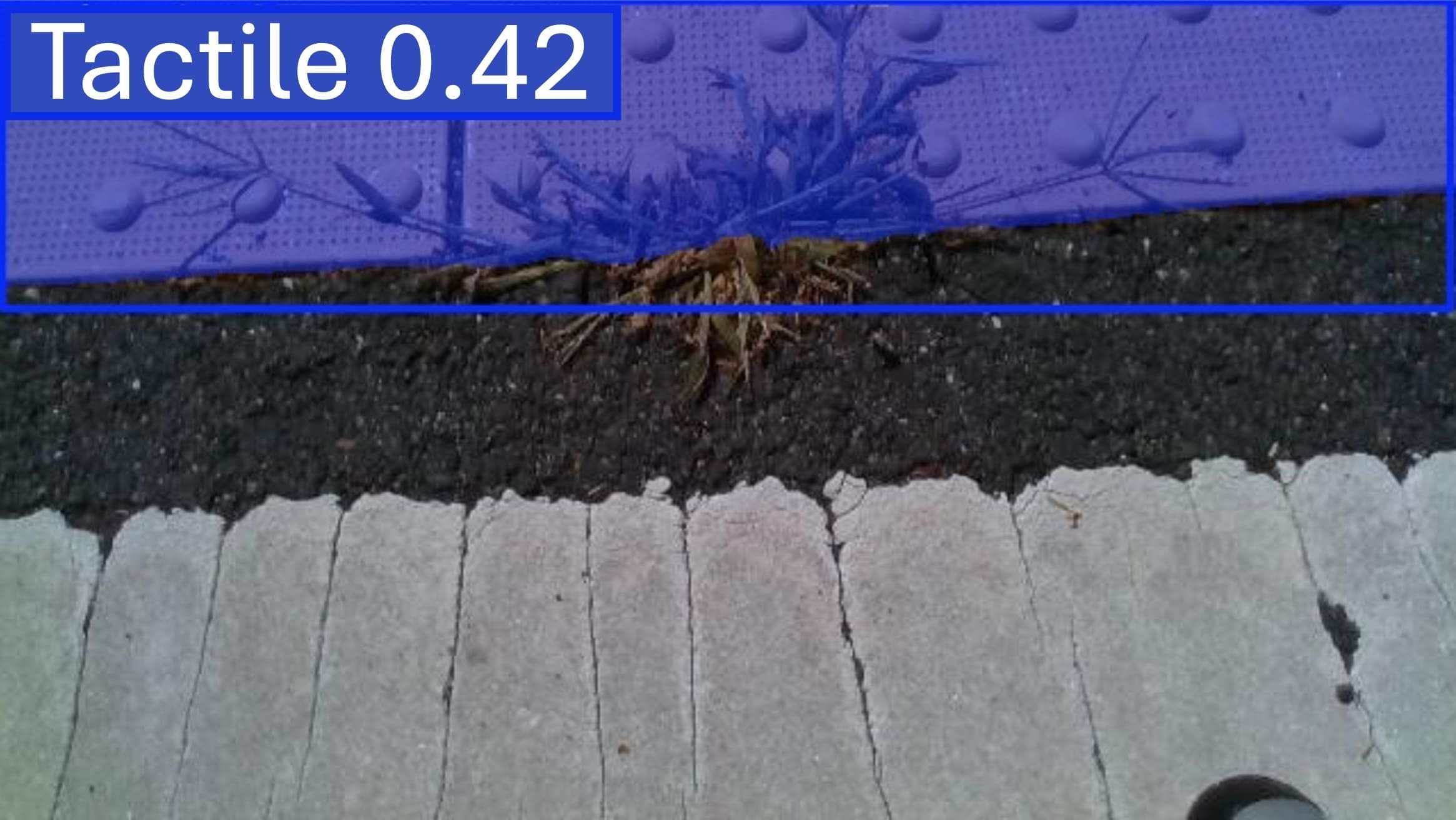} & \img{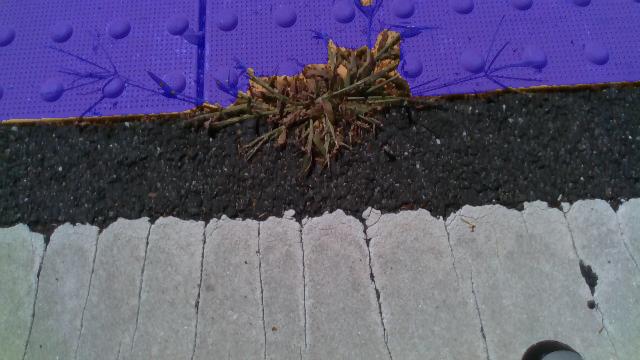}
           & \img{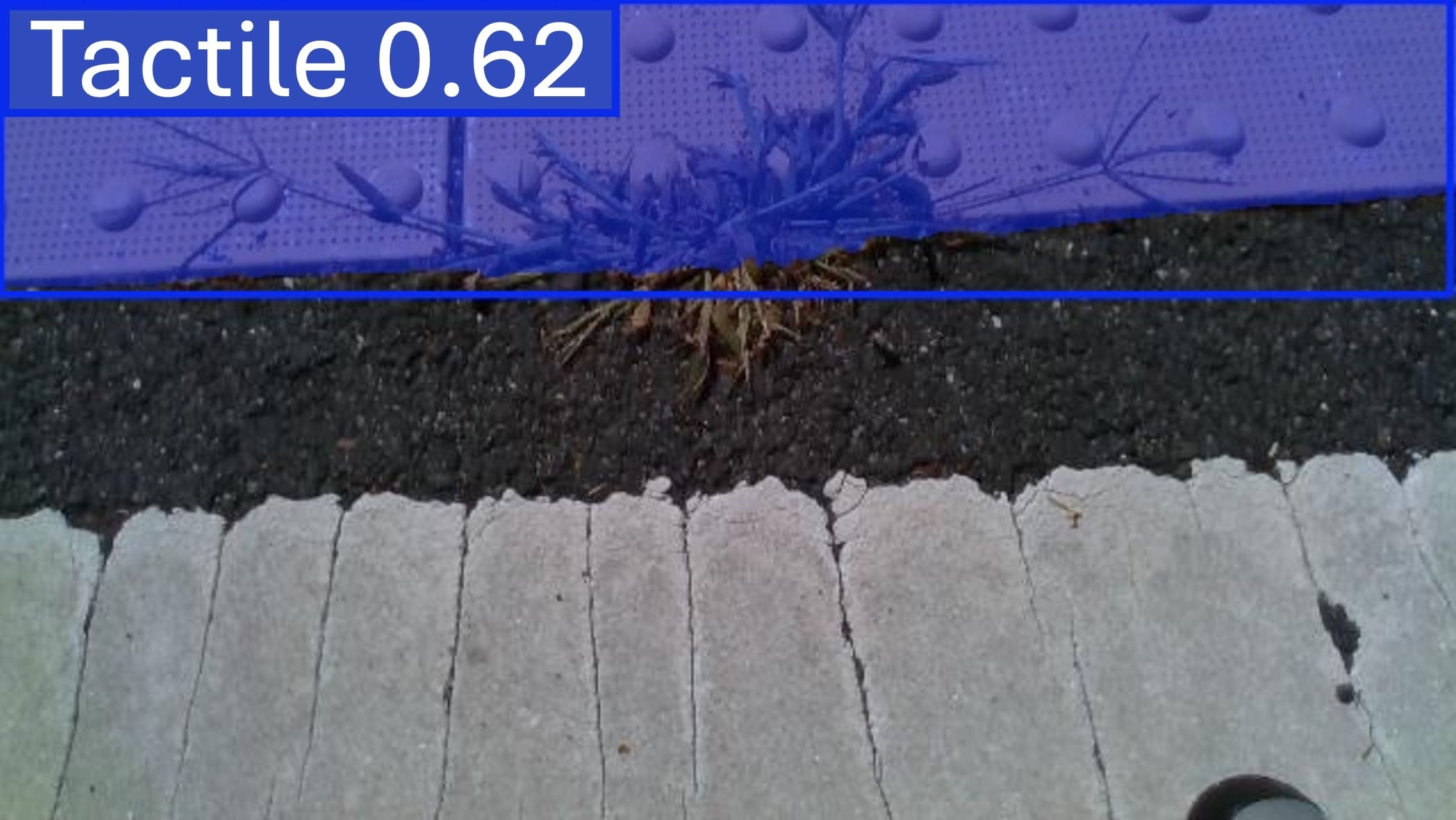} & \img{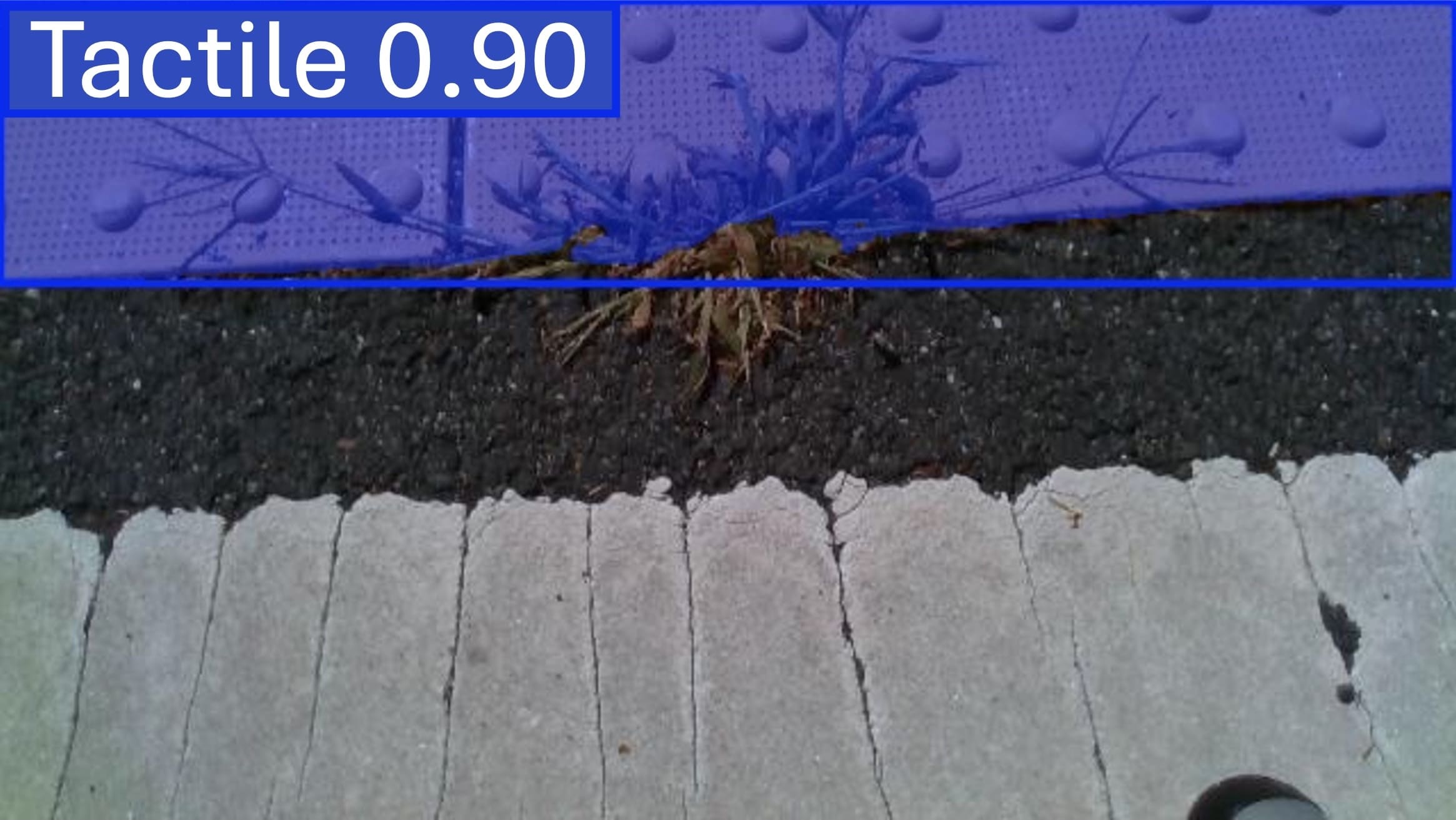} & \img{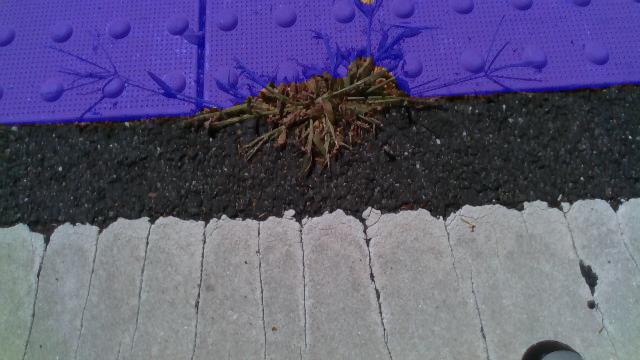} \\
   \footnotesize Sample
   & \footnotesize YOLOv11-seg-N & \footnotesize YOLOv11-seg-X & \footnotesize DINOv3+EoMT     
   & \footnotesize YOLOv11-seg-N& \footnotesize YOLOv11-seg-X& \footnotesize DINOv3+EoMT      \\
\end{tabular}
\end{adjustbox}
};

\draw[red, line width=0.9pt]
  ($(table.north east)+(-0.435\textwidth,-1.45em)$)
  rectangle
  ($(table.south east)+(-0.12,1.5em)$);

\end{tikzpicture}

\vspace{-3mm}
\caption{\textbf{Qualitative comparison of segmentation models trained on real data only vs. real + synthetic data.} 
Each row shows a test sample alongside predictions from YOLOv11-seg-N, YOLOv11-seg-X, and DINOv3+EOMT. Highlighted regions in the images show that models trained with synthetic data produce sharper boundaries and fewer missed detections, especially under challenging textures and lighting.}
\label{fig:qual}
\end{figure*}

\section{EXPERIMENTS}
\label{sec:experiments}

We assembled the \emph{GuideTWSI} dataset, comprising RBar-22K (21,994 real samples), SDome-15K (15,010 synthetic samples), and RDome-2K (2,466 real samples). To address three key research questions:
\begin{itemize}
\item \textbf{RQ1:} Is existing TWSI data sufficient for accurate truncated domes segmentation from a robot’s perspective?
\item \textbf{RQ2:} Can synthetic data augmentation improve segmentation performance?
\item \textbf{RQ3:} How reliably can a model fine-tuned on real and synthetic data perform on a real robot for stopping at truncated domes?
\end{itemize}
We used RBar-22K and SDome-15K for training and reserved RDome-2K for testing. As mentioned in Section~\ref{sec:realdata}, all data underwent rigorous preprocessing to ensure consistency. This process enabled experiments across several segmentation models fine-tuned either on real-only or synthetic-augmented training data. The training data were partitioned into training, validation, and test subsets using an 88\% / 6\% / 6\% split.

\subsection{Model training and setup}
We benchmark several segmentation models under identical training splits and evaluation metrics. Our comparison includes two variants of the latest YOLOv11 segmentation models (YOLOv11-seg-N and YOLOv11-seg-X~\cite{y11}), the transformer-based Mask2Former~\cite{cheng2022masked}, a SAM2.1+UNet model (comprising a SAM 2.1~\cite{ravi2024sam} backbone with a custom decoder for truncated dome class predictions), and two Vision Transformer (ViT)~\cite{dosovitskiy2020image} models fine-tuned from the recently released DINOv3-S~\cite{simeoni2025dinov3} (a patchwise regression classifier (RegCls) and the state-of-the-art Encoder-only Mask Transformer (EoMT) model~\cite{kerssies2025your}. 

\begin{figure}
    \centering
    \includegraphics[width=.9\linewidth]{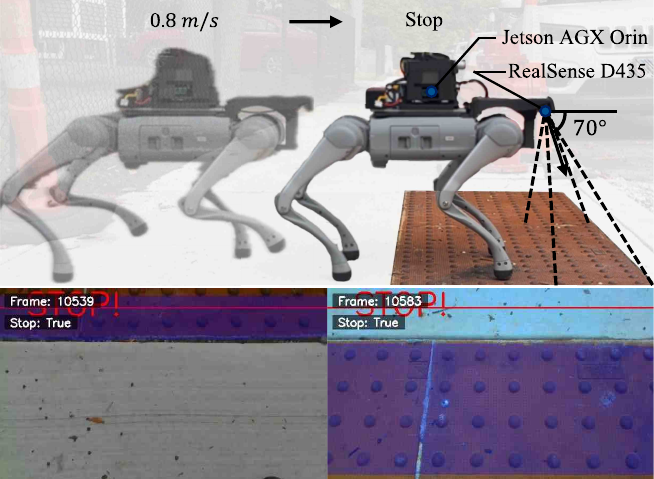}
    \caption{\textbf{Hardware configuration and segmentation visualization.} (Top) Hardware configuration with a downward-facing RGB camera mounted at a $70^{\circ}$ angle. (Bottom) Segmentation masks from the fine-tuned model trigger a stop command as the robot walks forward at \SI{0.8}{\meter\per\second}.}
    \label{fig:hw}
\end{figure}

We fine-tuned YOLOv11-Seg models using pre-trained weights with 640$\times$640 resolution using standard augmentations (e.g., mosaic cropping, flips, color jitter) for 100 epochs using the default stochastic gradient descent optimizer (initial learning rate of 0.01 and momentum 0.937) and a batch size of 16. Mask2Former~\cite{cheng2022masked} was trained using the standard Detectron2 configuration ($\approx 355\text{k}$ iterations) using the AdamW optimizer with $1024\times1024$ large-scale jittering augmentation crops, and automatic mixed precision being enabled. For the SAM2.1+UNet model, since SAM2.1 provides strong class-agnostic segmentation but lacks class awareness, we froze the SAM2.1 backbone and added lightweight decoder heads for supervised fine-tuning. Specifically, we designed a U-Net–style decoder that processes the $64\times64$ encoder features (256 channels) and progressively upsamples through four stages ($64\rightarrow 128\rightarrow 256\rightarrow 512\rightarrow 1024$), with skip connections at each scale. This enables dense, class-specific segmentation of truncated domes at the original image resolution. The decoder was trained for 100 epochs with AdamW optimizer with batch size of 4. Finally, the RegCls and EoMT models that use DINOv3 features were fine-tuned for 100k iterations. The EoMT model uses 100 learnable queries and an attention-mask annealing strategy, while RegCls classifies quantized patch embeddings into truncated dome vs. background with a regularization strength term. All segmentation models were evaluated on the RDome-2K real-world test set using standard segmentation metrics: precision, recall, mAP50–95, and mean Intersection-over-Union (mIoU).

\subsection{Experimental Results}

Using only the real data (left half of Table~\ref{fig:quant}), the models achieved moderate segmentation accuracy on truncated domes. The highest mIoU was 0.7389 (YOLOv11-seg-X), while several methods scored much lower (e.g. Mask2Former mIoU 0.5777, SAM2.1+UNet 0.4789). Precision was generally high (0.80–0.95), but recall was substantially lower (around 0.52–0.82), indicating that the models tend to be conservative: many truncated dome regions were missed. For example, Mask2Former achieved 0.946 precision but only 0.598 recall. These results suggest that compiled data (RBar) does not fully cover truncated dome appearances, so models trained on it cannot reliably segment domes. This confirms that existing real-world datasets lack sufficient diversity to capture truncated dome variability, in other words, being insufficient for the task of stopping at truncated domes.

Adding our synthetic SDome-15K significantly boosts segmentation performance. As shown in the ``real+synthetic'' setting (right half of Table~\ref{fig:quant}), all models improved on every metric. Notably, Mask2Former's mIoU rose from 0.5777 to 0.8375 (+0.2598) and recall jumped from 0.5975 to 0.8669. DINOv3+EoMT showed the largest gain, from 0.5804 to 0.8756 (+0.2954). Precision remained high in all cases (e.g., YOLOv11-seg-N from 0.7958 to 0.8718), while recall consistently increased. This indicates that models augmented with synthetic data during training detected more truncated dome pixels and produced more complete segmentation masks. In short, synthetic data augmentation substantially enhances segmentation of truncated domes, which can also be qualitatively visualized as in Fig.~\ref{fig:qual}.

Collectively, these results highlight synthetic data as an essential complement to curated real-world TWSI datasets, enhancing truncated dome segmentation and enabling reliable integration into safety-critical robotic systems.

\begin{table}[t]
\label{tab:robottest}
\centering
\caption{Robot stopping performance across different environments. }
\begin{tabular}{l@{\hspace{1em}}c@{\hspace{1em}}c@{\hspace{1em}}c}
\toprule
Environment & N & Mean Distance (cm) & Success Rate \\
\midrule
Urban     & 6 & $35.2 \pm 9.2$  & 29/30 \\
Urban2    & 5 & $47.1 \pm 15.6$ & 25/25 \\  
Suburban  & 6 & $38.2 \pm 14.4$ & 32/34 \\
Residential      & 3 & $34.6 \pm 14.2$ & 14/15 \\
\midrule
\textbf{Overall} & \textbf{20} & $\mathbf{39.0 \pm 14.3}$ & \textbf{100/104 (96.15\%)}  \\
\bottomrule
\end{tabular}
\end{table}


\section{ROBOT DEPLOYMENT}
\subsection{System Hardware Description}
We integrated our fine-tuned segmentation model into a fully untethered guide dog robot platform to assess real-time performance in a mobility assistance scenario. The system is built upon the Unitree Go2 quadruped robot equipped with a RealSense D435 camera positioned with a downward tilt for truncated dome segmentation as shown in Fig.~\ref{fig:hw}. We converted the model from PyTorch checkpoint into a TensorRT-optimized engine, which improved inference time up to 43 FPS. We used an NVIDIA Jetson AGX Orin computer for running inference of the finetuned YOLOv11-seg-N model (trained on RBar-22K + SDome-15K). 




\subsection{Precise stopping with segmentation masks}
Our system employs segmentation-based closest-point detection strategy, identifying the lowest truncated dome-pixel in the image frame as the point closest to the robot body. A stop command is triggered once this point exceeds a certain image height (e.g., image height × 0.1). This pixel-level method achieves a high stopping success rate while sustaining real-time performance on the Jetson AGX Orin, and it further offers the potential to convey orientation information to users through segmentation masks.


\subsection{Field trial results}
We tested the complete system in unseen real outdoor environments to validate its performance for stopping at truncated domes. The robot was taken to five different sites containing 21 truncated domes ranging from busy urban sidewalks to quiet residential streets (see Fig.~\ref{fig:realrobotmap} (b)). At each site, the robot was programmed to walk toward the truncated domes multiple times at \SI{0.8}{\meter\per\second}. We then recorded the success rate and whether it correctly stopped at an appropriate distance. Overall, the guide robot achieved a 96.15\% success rate (see Table~\ref{tab:robottest}). In all successful cases, the robot halted at a safe distance (approximately $39~\si{\centi\meter}$ from the starting point of the truncated dome and before the curb), which is sufficient space for a user to halt and prepare to step onto the crossing. We note that no false positive stops were observed – the robot never stopped erroneously when no truncated domes were present, thanks to the high precision of the segmentation. Failures were rare (4 out of 104 trials) and typically involved extremely challenging conditions, such as harsh lighting that caused severe lens flare. Even in these cases, the robot usually detected the tactile domes late (stopping a bit beyond the ideal point) rather than completely missing them.

\section{CONCLUSION}
Through insights from BLV guide dog handlers, trainers, and an O\&M specialist, we confirmed the critical importance of detecting truncated domes. Developing a reliable system to halt BLV travelers and notify them of safety-critical sites requires diverse, robot-relevant data---a resource largely missing from existing datasets. To address this gap, we introduced \emph{GuideTWSI}, a large-scale, diverse dataset for TWSI segmentation that combines photorealistic synthetic renderings, curated open-source annotations, and real-world quadruped collections. By addressing the geographic bias of prior datasets and the lack of robot-relevant viewpoints, GuideTWSI enables robust recognition of safety-critical truncated domes. Our experiments demonstrate that synthetic augmentation not only improves state-of-the-art segmentation accuracy but also directly enhances real-robot performance in safety-critical stop behaviors for BLV travelers.

\addtolength{\textheight}{-12cm}   






\section*{ACKNOWLEDGMENT}
The materials used in this study have been reviewed and approved by the university IRB (protocol IDs: 5709 and 6690). The study is supported by the National Institutes of Health (R21EY037411), the National Science Foundation (2427788), and the NVIDIA Academic Grant Program. 



\bibliographystyle{./IEEEtran} 
\bibliography{./IEEEabrv,./reference}

\end{document}